\documentclass{article}

\PassOptionsToPackage{numbers, compress}{natbib}

\usepackage[main, final]{neurips_2026}

\usepackage[utf8]{inputenc} 
\usepackage[T1]{fontenc}    
\usepackage{hyperref}       
\usepackage{url}            
\usepackage{booktabs}       
\usepackage{amsfonts}       
\usepackage{nicefrac}       
\usepackage{microtype}      
\usepackage{xcolor}         

\usepackage{graphicx}
\usepackage{subcaption}
\usepackage{multirow} 
\usepackage{amsmath}
\usepackage{amssymb}
\usepackage{mathtools}
\usepackage{amsthm}
\usepackage{xurl}
\usepackage{wrapfig}

\graphicspath{{figures/}}

\title{Discrete Diffusion Models Exploit Asymmetry\\to Solve Lookahead Planning Tasks}

\author{%
    Itamar Trainin \\
    Hebrew University of Jerusalem \\
    \texttt{itamar.trainin@mail.huji.ac.il} \\
    \And
    Shauli Ravfogel \\
    New York University \\
    \texttt{shauli.ravfogel@nyu.edu} \\
    \And
    Omri Abend \\
    Hebrew University of Jerusalem \\
    \texttt{omri.abend@mail.huji.ac.il} \\
    \And
    Amir Feder \\
    Hebrew University of Jerusalem \\
    \texttt{amir.feder@mail.huji.ac.il} \\
}

\begin{document}
\raggedbottom
\maketitle

\begin{abstract}

Discrete diffusion models are often said to outperform autoregressive models on planning tasks, but the specific mechanisms enabling this are not yet well understood. To explain why, we abstract the complexity of studying ``planning'' into graph traversal problems, which offer a minimal setup for evaluating language models' planning and lookahead capabilities. We find that non-autoregressive language models are able to leverage an inherent directional asymmetry in lookahead planning tasks to outperform the training efficiency of their autoregressive counterparts. Through a mechanistic study of the training and inference dynamics of autoregressive and non-autoregressive models, we discover that while forward traversal through junctions in the graph requires complex sequential planning, the reverse path remains deterministic. Consequently, non-autoregressive models naturally develop a fundamentally different solution strategy. Effectively, they learn a reverse-decoding pattern which reduces an $\ell^{th}$-order lookahead problem to a sequence of simple $1^{st}$-order transitions. Latent-space analysis further confirms these divergent strategies by demonstrating the distinct internal representations formed during generation. Ultimately, these findings highlight the advantages of non-autoregressive modeling while clarifying the inherent planning limitations induced by autoregressive architectures.

\end{abstract}


\section{Introduction} \label{sec:intro}

\begin{figure}[!htbp]
    \centering
    \includegraphics[width=\linewidth]{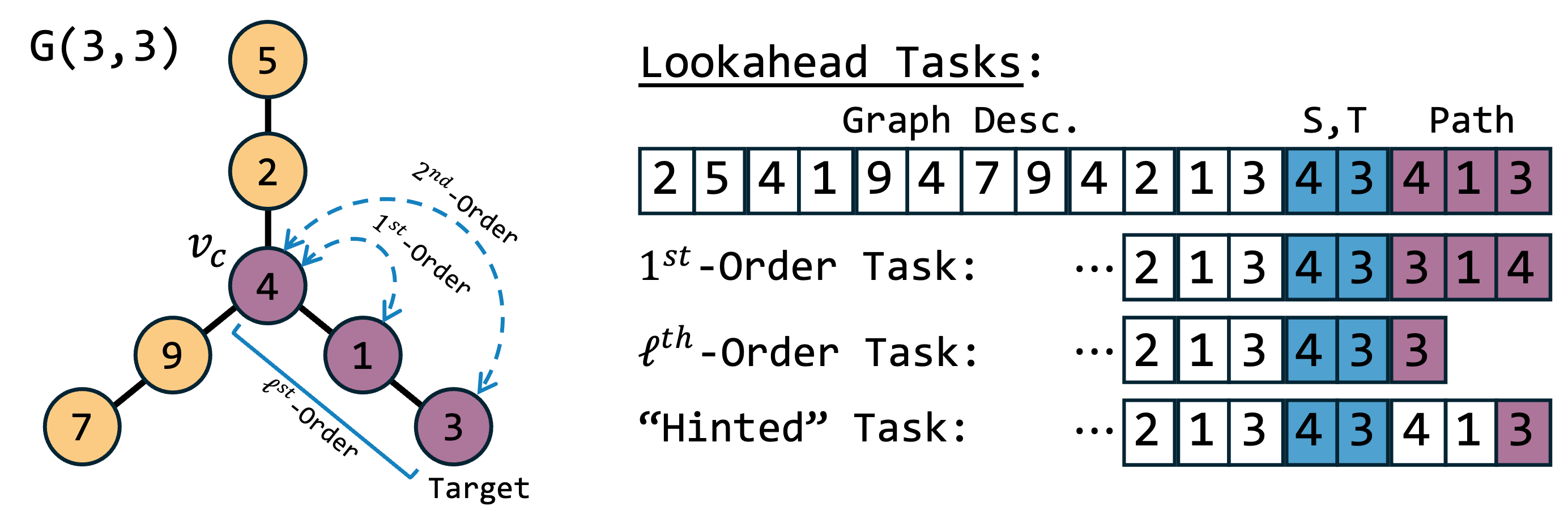}
    \caption{A Star-Path task graph ($d{=}3$ arms, $l{=}3$ vertices per arm), and its equivlent string representation for \textit{Original} lookahead task, \textit{$1^{st}$-Order} task where the path is decoded in reverse, \textit{$\ell^{th}$-Order} task where only the first and second vertices are predicted, and \textit{``Hinted''} task where the true first and second vertices are provided in context. Purple indicates tokens predicted at test time.}
    \label{fig:star_graph}
\end{figure}

The ability to plan a valid path towards reaching a desired goal is a fundamental requirement for complex decision-based systems \cite{valmeekam2023planning}. To successfully plan, an agent must maintain a model of potential state transitions, looking ahead to understand not just immediate effects but how current choices constrain future possibilities \cite{gao2024large, erdogan2025plan}. This lookahead often creates a structural asymmetry. The forward planning requires resolving complex ambiguities at branching junctions where multiple valid choices diverge, while the reverse process of tracing from a known goal back to the start is junction-free and thus deterministic (see Fig. \ref{fig:star_graph}). 

Consider an itinerary planning task where an agent must schedule stops between two cities. Planning in the forward direction means that choosing a stop along the way requires evaluating how that decision restricts all subsequent options. Therefore, an agent must compute the entire trajectory, ensuring the final destination remains reachable. Conversely, starting from the goal and working backward can dissolve this complexity. Although Autoregressive (AR) modeling is the current standard for Large Language Models (LLMs), a recent line of work suggests that the causal ``next-token'' training objective poses an inherent conflict with this lookahead requirement \cite{bachmann2024pitfalls, nie2025large, ye2025dream, ye2024beyond}.

Limited by their causal decoding direction, AR sequential models are unable to exploit the asymmetry in the lookahead task, forcing such model families to commit to a decision at a junction before the goal is fully processed. To mitigate this, AR models often rely on generating intermediate reasoning steps, effectively using their generated context as a ``scratchpad'' \citep{nye2021show} to explicitly compute the necessary trajectory before yielding a final answer. However, generating these intermediate steps is not an out-of-the-box capability inherent to the sequential architecture itself, but rather relies on dedicated training. Consequently, \citep{bachmann2024pitfalls} confirmed empirically that without such explicit intermediate scaffolding, AR models, whether Transformer-based \citep{vaswani2017attention} or State-Space \citep{gu2024mamba, dao2024transformers}, default to random guessing at critical junction vertices, yet succeed when the task is reversed (junction-free). This fundamental limitation motivates our investigation of whether Non-Autoregressive (NAR) generative models offer a more suitable paradigm.
            
We thus hypothesize that NAR models can reduce complex long-distance lookahead to $1^{st}$-order neighbor transitions by exploiting the asymmetry in directionality. We focus on Discrete Diffusion Models (dLLMs) \citep{campbell2022continuous, campbell2024generative}, a class of NAR models that generate complete sequences by iteratively refining from noise. The training paradigm of dLLMs forces them to learn a more complex class of problems \citep{kim2025train}, potentially improving their effectiveness on benchmarks that require planning abilities, as has been recently shown by \cite{ye2024beyond, nie2025large, ye2025dream}.

Rigorously assessing planning and lookahead capabilities in LLMs is challenging due to their ill-defined nature and the inherent lack of experimental control when evaluating them at full scale \citep{valmeekam2023planning}. Nonetheless, one promising framework is the Star-Path task \citep{bachmann2024pitfalls}, which gives a controlled perspective of planning as a graph-traversal problem. In this task, the model receives a textual graph description along with source and target vertices, and must produce the connecting path. By introducing junctions to the graph topology,  the model cannot determine the correct arm from local information alone. As such, the task enables a controlled planning setup in which the mechanisms learned by each architecture can be isolated.

To explore this hypothesis, we present a study extending the Star-Path framework to observe how these two generative paradigms tackle graph traversal (see Fig.\ref{fig:star_graph}). Our investigation reveals a striking disparity in sample efficiency, as graph complexity scales, AR models require exponentially more data to converge than their NAR counterparts. We conduct a mechanistic study to identify the reason for this fundamental divergence in learning. While AR models hit a bottleneck attempting to master $\ell^{th}$-order lookahead and build global structural ``depth-maps'', NAR models exploit structural asymmetry through a reverse-decoding strategy. By working backward from the goal, they reduce the planning problem to simple $1^{st}$-order transitions and rely on local identity matching. Ultimately, this study demonstrates how the flexible bidirectional context of Discrete Diffusion models enables solution strategies that are not only fundamentally different but also efficient for lookahead planning\footnote{Code and data available on: https://github.com/itamartrainin/dllms-planning.git}.


\section{Related Work} \label{sec:related_work}

\subsection{Star-Path: A Textual Graph Algorithm Problem} \label{sec:star_path}

The Star-Path task represents a key subset of broader research evaluating LLMs through textually represented graphs \citep{jin2024llmgraphs, ye2024languagegraph, wang2023lmgraphproblems, fatemi2023talklikeagraph}. 
Specifically, it aims to isolate planning and reasoning mechanisms via a synthetic and controllable environment. 
Distinguished by its minimality, the task has emerged as a benchmark for highlighting the limitations of current architectures in sequential processing \citep{bachmann2024pitfalls, saparov2025transformers, sanford2024understanding, noroozizadeh2025deep, zhang2022unveilingtransformerslego, frydenlund2025language}.

Initially, \cite{bachmann2024pitfalls} used the Star-Path task to suggest that autoregressive (AR) architectures are fundamentally unable to solve lookahead tasks. Yet, subsequent work has demonstrated that the task is nonetheless solvable using such architectures \citep{frydenlund2024pathologicalpathstar, noroozizadeh2025deep, frydenlund2025language}. 
Other studies have utilized the Star-Path setting to investigate the learnability of search versus retrieval \citep{saparov2025transformers}, the implicit encoding of geometrical relationships \citep{noroozizadeh2025deep}, and the computational complexity hierarchy of graph reasoning \citep{sanford2024understanding}.


\subsection{Non-Autoregressive Language Models} \label{sec:nar_models}

While the field has largely converged on AR models due to their simplicity and scalability \cite{radford2018improving, radford2019language, brown2020language}, NAR paradigms have re-emerged as a compelling alternative \cite{devlin2019bert, raffel2020exploring}. 
Most notable among these are Discrete Diffusion Language Models (dLLMs) \citep{campbell2022continuous, campbell2024generative}, which, unlike earlier NAR architectures, combine bidirectional context with robust generative capabilities. 
Recent scaling efforts demonstrate that dLLMs can achieve performance competitive with state-of-the-art AR baselines at comparable costs \cite{gat2024discrete, nie2025large, ye2025dream}.

Operating under a distinct training-inference paradigm, dLLMs forgo a single left-to-right factorization. Instead, they learn to denoise partially corrupted sequences in parallel, capturing a broad family of conditional distributions. Generation is then facilitated through an iterative denoising process that progressively reconstructs the sequence from noise \cite{gat2025set, kim2025train, ben2025accelerated}. Consequently, dLLMs exhibit superior performance on complex planning tasks \cite{ye2024beyond}. Yet, the underlying mechanisms that drive this remain largely underexplored.


\section{Method} \label{sec:method}

To overcome the challenges of evaluating lookahead planning in full-scale language models, we utilize a minimal setup where the lookahead mechanism can be studied under control. Consider a graph $G:=(V,E)$ and two  vertices $v_c, v_t\in V$. Identifying a valid path traversing between these vertices requires a fundamental understanding of the graph's structure, including the possible transitions and a strategy for composing them successfully. Building upon recent research employing graph structures to analyze the emergence of algorithmic solutions during LLM training \citep{bachmann2024pitfalls, sanford2024understanding, saparov2025transformers, noroozizadeh2025deep}, we leverage this setup to compare how AR and NAR models develop lookahead mechanisms by introducing junctions that obscure the direct path between vertices. 

We train AR and NAR models by presenting them with arbitrary graph structures containing branching junctions, along with source and target vertices. The models are tasked with predicting the solution as a sequence of vertices forming the connecting path. By fully randomizing both the graph structure and the source-target pairs, we ensure the models cannot rely on memorization, forcing them to learn generalizable traversal strategies. Representing these graphs as tokenized strings maps the traversal problem into the sequential domain, providing an ideal testbed for investigating how different decoding paradigms handle long-distance dependencies and lookahead requirements. While many graph structures comply with this broad definition, we simplify our experimental setting to a star-shaped graph \citep{bachmann2024pitfalls}, a minimal graph structure containing a single decision point. This reduces the task to a precise decision point, allowing us to study the learned lookahead mechanisms in isolation. This setup naturally scales, as any junction-based graph can be represented as a composition of star graphs. Consequently, our primary investigation focuses on the star-graph setting, while more intricate composite structures are explored in Appendix \ref{app:two_sided_star}.


\subsection{The Star-Path Task} \label{sec:the_star_path_task}

A single Star-Path problem (an example in the data) is defined as a triplet consisting of a \textit{graph description}, a \textit{source-target vertex pair}, and the \textit{unique connecting path}. As depicted in Fig.\ref{fig:star_graph}, a Star-Graph ${G(d,l):=(V_G,E_G)}$ is a graph whose vertices ($V_G$) and edges ($E_G$) form a star configuration, featuring $d\in\mathbb{N}$ arms extending from a single center vertex $v_c\in\mathbb{V}$, each of length $l$. Consequently, the set of vertices is defined to be $V_G:=\{v_c, v_{1,1},...,v_{1,l-1},...,v_{d-1,l-1}\}\subseteq\mathbb{V}$ so that ${|V_G|=1 + d(l-1)}$. We define $\mathbb{V}$ as the global pool of allocatable vertices such that $|\mathbb{V}|=N$. The vertices on the graph are connected into a star shape by the set of edges $E_G:= \left\{\, (v_{i,j-1}, v_{i,j}) \;\middle|\; i \in [0,d),\; j \in [1,l), v_{\cdot,0}:= v_c,\; v_{i,j} \in V_G \right\}$. The source-target pair is defined as $\mathcal{T}:=(v_c,v_{t,l-1})$, where $v_{t,l-1}$ is a leaf vertex terminating the $t$-th arm $t\in[0,d)$. Finally, the connecting path is defined to be the ordered (inclusive) sequence of vertices from the source to the target $\mathcal{P}:=(v_{t, i}:i\in[1,l);v_{t,0}:=v_c)$.

For fixed star parameters $(d,l)$, the $i$-th datapoint is defined textually as the triplet $S:=[E_{G_i}\mathbin{\|}\mathcal{T}_i\mathbin{\|}\mathcal{P}_i]$. The graph $G_i(d,l)$ is constructed by uniformly sampling vertices without replacement from $\mathbb{V}$, yielding an overall sequence length of $|S|=(2\cdot|E_{G_i}|) + 2 + (d(l-1) + 1)$. To minimize confounding factors, we do not introduce separator tokens to mark sequence attributes (e.g., boundaries between the edge list, the source–target pair, and the path). Within this setup, because the source vertex $v_c$ is connected to $d$ identical arms, the correct first transition to $v_{t,1}$ is highly ambiguous. To generate this single token, an autoregressive model must identify the target vertex $v_{t,l-1}$ within the provided graph structure, recognize which arm $t$ it belongs to, and successfully uncover the $l-1$ edges back to the center. Thus, a seemingly simple next-token prediction step inherently requires an $l$-th order lookahead operation.


\subsection{Variations of the Star-Path Task} \label{sec:variations}

To fully study the lookahead mechanism, we introduce three additional variations to the original Star-Path task, controllably isolating different traits of the task. All variations are depicted in Fig.\ref{fig:star_graph}.

\textbf{$1^{st}$-Order Lookahead Task}: Models are trained to generate the ``reverse-path'' leading from the leaf vertex back to the center vertex. Since the reverse direction is junction-free, only first-order transitions need to be acquired by the model.

\textbf{$\ell^{th}$-Order Task}: Models are trained to predict only the second vertex in the forward path, which is the critical junction choice. Under this setting, the vertices on the path from the leaf backwards cannot be used by the model. This restriction prevents models from using backward tokens as a scratchpad.

\textbf{``Hinted Task''}: Like the original task, models are trained to predict the forward-going path, but they are hinted with the first and second vertices. This way, the critical junction choice is eliminated, requiring the modeling of only first-order transitions.


\subsection{Experimental Setup} \label{sec:experimental_setup}

A generative model $LM_\theta(\boldsymbol{x}_{\text{prefix}})=P(\boldsymbol{x}_{\text{path}}|\boldsymbol{x}_{\text{prefix}})$ is trained to solve the Star-Path task as defined in \S\ref{sec:the_star_path_task}, where the task sequence is provided as the \textit{prefix} and the \textit{path} is the predicted sequence representing the solution. To compare the behavior of AR and NAR models, we train equivalent models. For the NAR variant, we adopt the Discrete Diffusion modeling paradigm first introduced by \citep{gat2024discrete}. To ensure equivalence, we use a shared transformer backbone based on the \texttt{DDiT} implementation described by \citep{gat2024discrete, lipman2024flow, ben2025accelerated}. For AR modeling, we modify this implementation to (1) use causal self-attention, (2) set the diffusion time embedding to a constant value $t=0$, and (3) replace the diffusion training and sampling procedures with standard AR next-token training and decoding.

In our experiments, we utilize a decoder-only transformer architecture \citep{vaswani2017attention} scaled to match standard small LLMs. This design choice facilitates direct comparisons with real-world models while maintaining computational efficiency during training. The vocabulary is defined to be the enumeration of $\mathbb{V}$, alongside a \texttt{[MASK]} token. All sequences within each training setup are of a fixed length. Refer to appendix \S\ref{app:architecture} for model specifications and computational setup.

For a given experiment, we fix graph parameters (e.g., $d$,$l$) and construct a single held-out test set consisting of $20K$ unique Star-Path problems, which is shared across all runs. To support extreme training set size, the training examples are generated on-the-fly at train time by sampling from a generator function. We ensure no sample repetition and no overlap with the test set. For each run, we record the number of training examples required to reach convergence, and enforce a maximum training budget after which training is terminated. Because training samples are generated online, no two runs observe the same samples or sample order, which further reduces potential confounding effects.


\subsection{Training paradigms} \label{sec:training_paradigms}

As we are interested in generating the solution sequence given the task prefix, two possible training regimes can be employed. In a \textit{conditional training} approach, the model is conditioned on the prefix sequence and trained exclusively on the path tokens. Consequently, the loss on the prefix tokens is masked out, and gradients are computed only for the path tokens. Alternatively, in a \textit{full-sequence training} approach, the model is trained to generate the entire sequence (both prefix and path). Here, gradients are computed over all tokens, additionally rewarding the model for learning to generate valid graph descriptions. This method closely resembles the pre-training process used for standard LLMs.

In this work, we use the full-sequence training regime, as we found that conditional training requires a significantly larger number of samples to reach convergence. We provide a detailed comparison of the two regimes in Appendix \S\ref{app:learnability}, including ablation results demonstrating that prefix gradients are critical for AR convergence. Across all experiments, at test time, we provide the ground-truth prefix and decode only the path tokens.


\subsection{Success Criteria} \label{sec:evaluation}

While standard loss and perplexity criteria are used to train the model, our primary focus is to evaluate performance on the lookahead task itself. For this reason, we measure convergence using an \textit{exact-match} metric, formally defined in Appendix \ref{app:evaluation}. A prediction is considered correct only if the entire predicted path matches the ground-truth path token-by-token. A model is considered to have converged when accuracy exceeds $>0.95$ for $t=10$ consecutive validation steps, or when the maximum budget is reached.


\section{Findings} \label{sec:findings}

\subsection{Discrepancy in Convergence Complexity} \label{sec:convergence}

\begin{figure}[!h]
    \centering
    \includegraphics[width=0.6\textwidth]{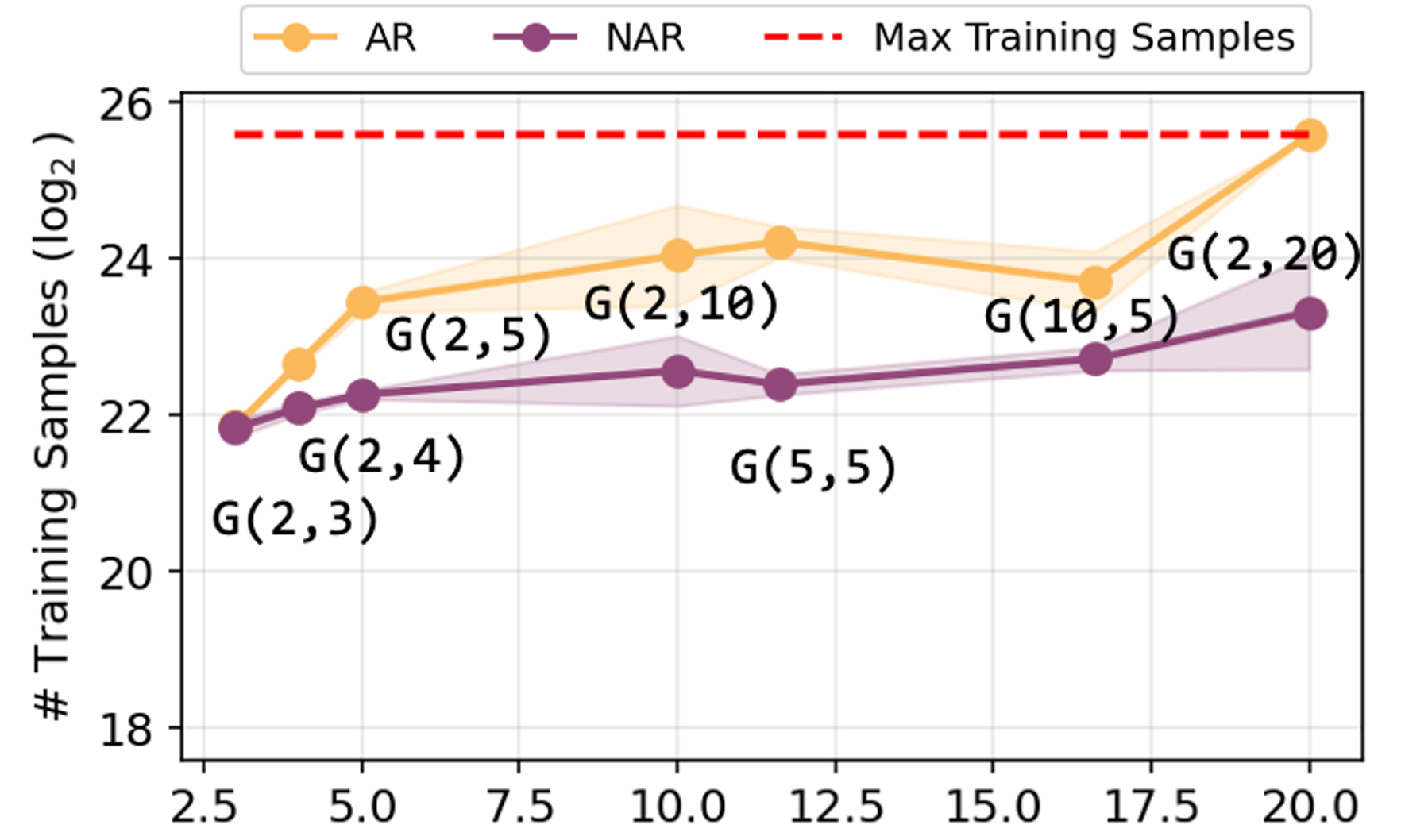}
    \caption{Comparison of AR and NAR convergence complexity as a function of Star-Path graph size. Axes ($log_2$ scale) show associative connections ($x$) and samples to convergence ($y$) when training AR and NAR models on the Star-Path Task. The red line (50M samples) marks the termination threshold for inconclusive convergence. Shaded areas indicate variance across 10 seeded runs.}
    \label{fig:underlying_mechanism_a}
\end{figure}

Motivated by the asymmetry in the Star-Path task, we establish a baseline understanding of how AR and NAR models handle lookahead planning by evaluating their convergence behavior on the task. The results are reported in Fig.\ref{fig:underlying_mechanism_a}. 
Although they are trained the same and are able to achieve comparable performance, their sample complexity exhibits markedly different scaling behavior. For AR models, the required number of training samples grows rapidly. 
Conversely, NAR models exhibit a significantly flatter scaling trajectory. As the number of associative connections increases, the sample complexity gap between the paradigms widens substantially. Importantly, this discrepancy is robust to model scale, where significantly smaller and larger model sizes are evaluated (see  Appendix~\ref{app:model_scale}).

This naturally prompts the central hypothesis of our study of how NAR models learn to solve high-order planning lookahead tasks more efficiently. By attributing an associative memory structure to transformers \cite{bietti2023birth, radhakrishnan2020overparameterized}, we hypothesize that models learn to convert graph configurations into canonical representations for transition retrieval. The forward path from $v_c$ to $v_{t,l-1}$ requires $\ell^{th}$-order lookahead, but the reverse path is deterministic. NAR models can therefore adopt a reverse strategy requiring only $1^{st}$-order transitions. Since mastering higher-order transitions presupposes mastering simpler ones first, AR models face a compounding learning challenge that accounts for their steeper sample complexity. 

To assess whether this advantage extends beyond the controlled Star-Path setting, we additionally evaluate full-scale NAR (dLLMs) and AR LLMs on established planning benchmarks, showing the NAR's superiority on such tasks and how scratchpad improves AR performance (Appendix~\ref{app:real_planning}).


\subsection{Mechanistic Adaptability} \label{sec:adaptability}

\begin{figure}[!ht]
    \centering
    \includegraphics[width=\linewidth]{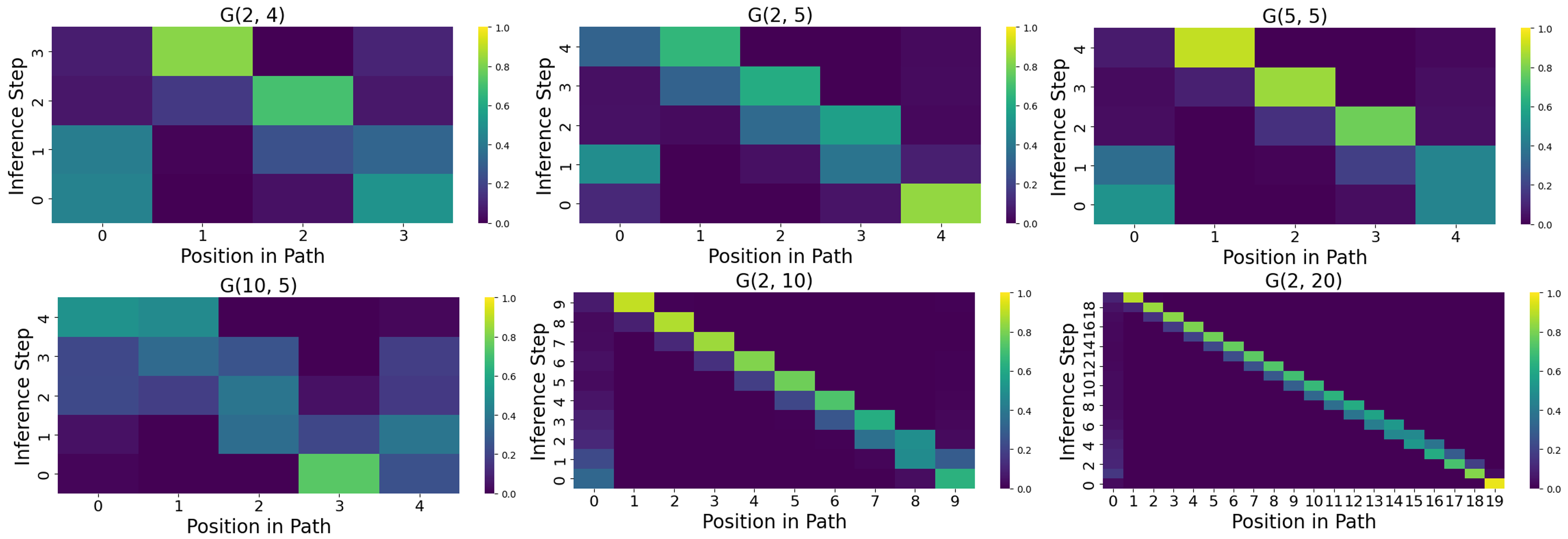}
    \caption{NAR models unmask tokens from goal to start, confirming reverse decoding. Each panel shows a different graph setting where the $x$-axis represents the vertex position in the path, and the $y$-axis represents the decoding step. Color intensity indicates the percentage of test examples (out of $20K$) where a token was unmasked at a given step.}
    \label{fig:nar_inference_dynamics}
\end{figure}

To validate the emergence of the reverse-decoding pattern in NAR models, we begin by investigating whether NAR models are capable of utilizing a flexible decoding strategy. While AR models are strictly restricted to the causal decoding direction, NAR models may implement alternative approaches. Fig. \ref{fig:nar_inference_dynamics} visualizes the decoding process of our NAR model (dLLM; see \S\ref{sec:experimental_setup}) for a common setup of single-token unmasking at every inference step. Across multiple graph settings, we observe a consistent diagonal pattern moving from the bottom-right to the top-left, confirming that the NAR model adopts the reverse decoding strategy as hypothesized.

\begin{figure}[!ht]
    \centering
    \includegraphics[width=0.7\textwidth]{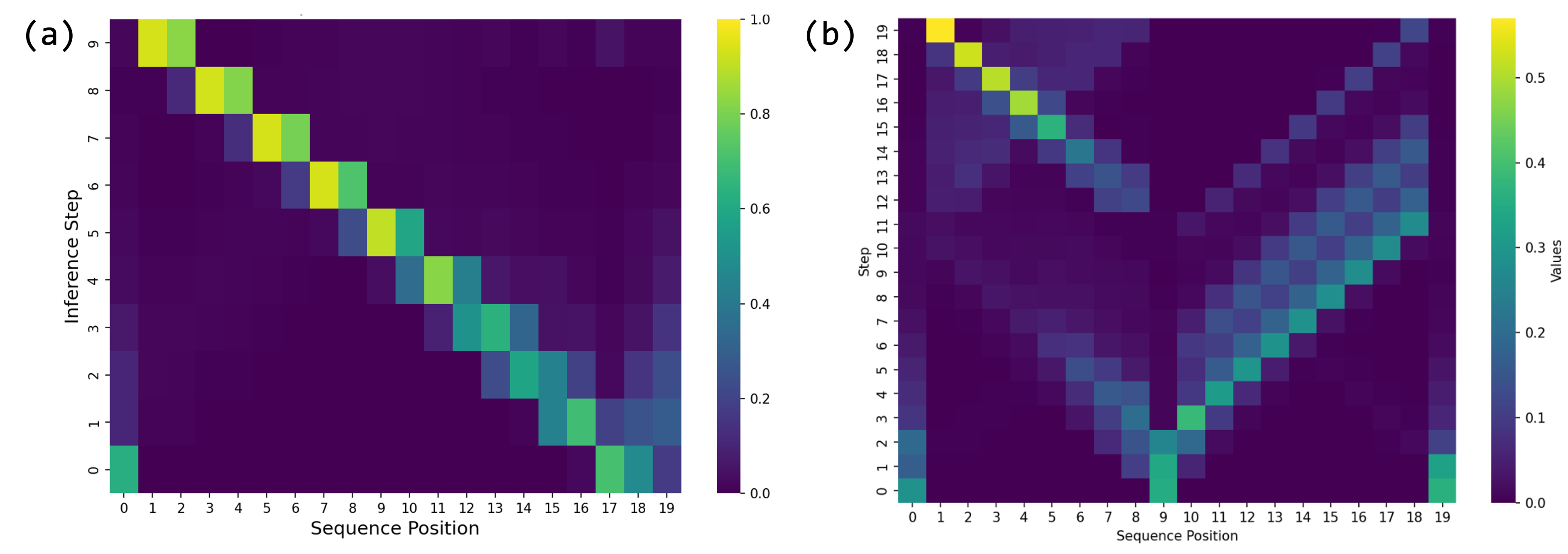}
    \caption{Decoding dynamics on the $G(2,20)$ graph for (a) unmasking of $k=2$ tokens at each decoding step and (b) NAR model (dLLM) trained on the Two-Sided-Star task. Color intensity indicates the percentage of test examples (out of $20K$) where a token was unmasked at a given step.}
    \label{fig:mech_additional_main}
\end{figure}

Crucially, this adaptable behavior generalizes to both accelerated sampling, block-wise decoding, and more complex graph topologies, where the NAR model successfully identifies deterministic anchors to bypass complex planning. Fig. \ref{fig:mech_additional_main}(a) shows the unmasking of multiple tokens simultaneously for faster decoding (top-$k=2,4,8$ in Appendix \ref{app:mech_additional:acce}), where the model consistently utilizes a reverse-decoding pattern. Furthermore, when tested on the alternative ``Two-Sided-Star'' task (see task formulation in Appendix \ref{app:two_sided_star}), a topology designed with junctions at both the source and target, the NAR model dynamically expands outwards from a deterministic mid-path vertex in both directions (Fig. \ref{fig:mech_additional_main}(b)). We further test this adaptability under block-wise generation; see Appendix \ref{app:mech_additional:block}, Fig.~\ref{fig:blockwise_decoding}.

\subsection{Exposing the Discrepancy in the Degree of Learned Transitions} \label{sec:exposing}

Building upon this observed adaptive behavior, we design further experiments to explicitly expose the fundamental discrepancy between the complex $l^{th}$-order and simple $1^{st}$-order transitions acquired by AR and NAR models. By introducing the targeted modifications defined in \S\ref{sec:variations} to the original Star-Path task, we control the specific types of transitions the models must acquire.

\begin{figure}[!h]
    \centering
    \includegraphics[width=\textwidth]{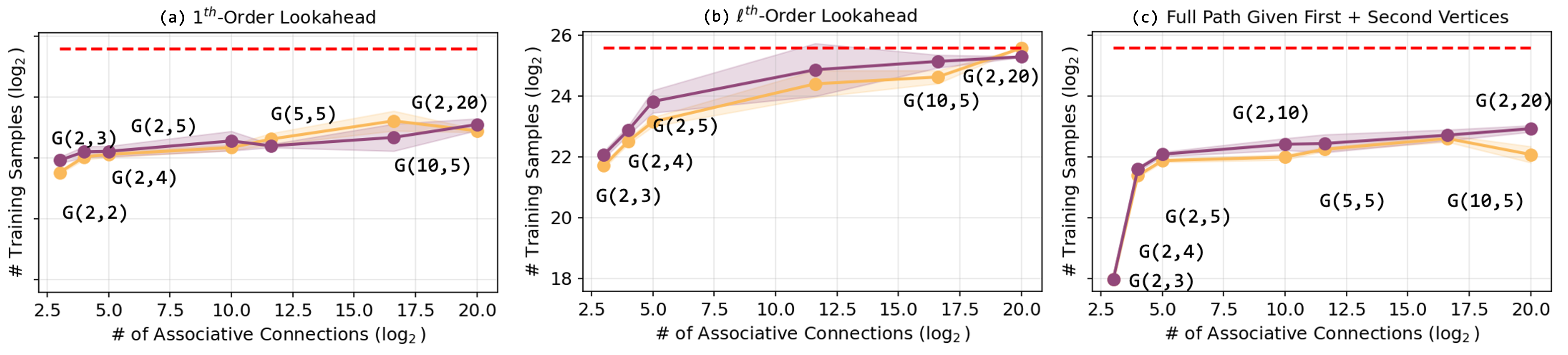}
    \caption{Comparison of $1^{st}$ vs. $\ell^{th}$-order lookahead across graph sizes. Axes ($log_2$ scale) show associative connections ($x$) and samples to convergence ($y$). Subplots represent: (a) $1^{st}$-Order, (b) $\ell^{th}$-Order, and (c) Hinted tasks. The red line (50M samples) marks the termination threshold for inconclusive convergence. Shaded areas indicate variance across 10 seeded runs.}
    \label{fig:underlying_mechanism}
\end{figure}

Fig. \ref{fig:underlying_mechanism}(a-c) compares sample complexity across these variations. The $1^{st}$-order task (a) enables AR models to use simple transitions, the $\ell^{th}$-order task (b) forces NAR models to learn complex transitions, and the Hinted task (c) eliminates the junction decision for both. When both paradigms must learn the same transition order, their sample complexities match, confirming the impact of the order of transitions that need to be learned on the sample complexity. Furthermore, the discrepancy thus originates from the causal constraint forcing AR models to learn $\ell^{th}$-order transitions.

\subsection{Hierarchy in Mastering Transitions}

Once established that AR models must learn complex $\ell^{\text{th}}$-order transitions, whereas NAR models only require simpler $1^{\text{st}}$-order transitions, we now deepen our investigation to understand why mastering these higher-order transitions demands a significantly larger sample complexity. Following the conclusions in \citep{sanford2024understanding}, we claim that a transformer model must master the simpler lower-order transitions first, using them to uncover higher-order ones, thereby creating a learning hierarchy in the process of learning to lookahead. To demonstrate this, we focus on the training dynamics.

\begin{figure}[!h]
    \centering
    \includegraphics[width=0.8\textwidth]{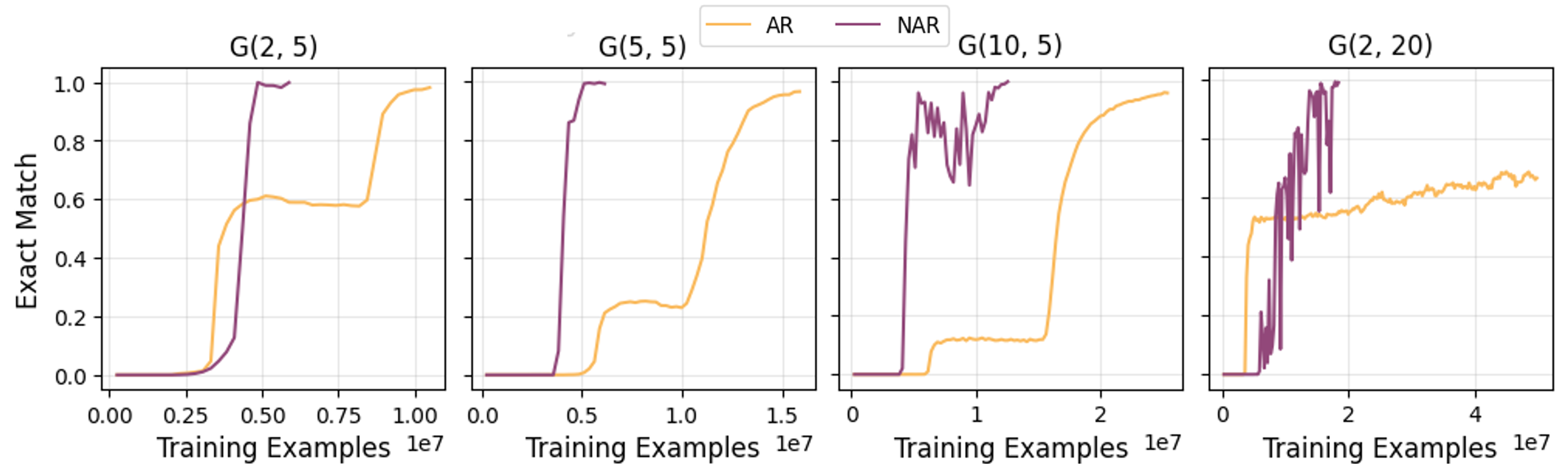}
    \caption{Train dynamics across varying graph configurations. AR models exhibit a two-phase convergence with a prolonged plateau, while NAR models converge in a single jump.}
    \label{fig:convergence_comparison}
\end{figure}

We show this hierarchical learning process by comparing the convergence dynamics. Fig. \ref{fig:convergence_comparison} depicts the \textit{exact-match} training accuracy at equidistant intervals during the training process. The resulting learning trajectories reveal a stark contrast. While NAR models exhibit a single, rapid transition to perfect accuracy, AR models undergo two distinct ``jumps'' before converging. Crucially, the first AR jump coincides exactly with the NAR model's single jump to convergence. 
This phenomenon provides empirical support for our conjecture. We posit that this initial jump corresponds to the point at which both models successfully internalize the simpler $1^{\text{st}}$-order transitions. 
Because NAR models rely exclusively on these local mappings, this initial mastery is sufficient for immediate convergence. 
In contrast, AR models must utilize these $1^{\text{st}}$-order transitions as a foundational prerequisite for deducing the more complex $\ell^{\text{th}}$-order relations. Consequently, the AR models enter a prolonged plateau phase to internalize these higher-order dependencies. This extended phase directly accounts for their larger sample complexity, culminating in a second jump to convergence only once the entire hierarchy of transitions is mastered. 
In Appendix \ref{app:training_dynamics}, we examine the convergence of individual token positions, where the same behavior is observed. This qualitative learning hierarchy is also robust to model scale, as increasing model capacity shifts the transition points but does not eliminate the separation between the early acquisition of simpler transitions and the
later AR convergence (Appendix~\ref{app:model_scale}). Furthermore, in Appendix \ref{app:mamba}, we experiment with a Mamba-based AR model, which exhibits smoother convergence and stronger encoding of graph structure while still showing a sample-efficiency gap relative to NAR.

\subsection{Latent Space Implications} \label{sec:embeds}

To further substantiate our results, we investigate the latent space representations learned by both architectures. Our findings indicate that solving the task requires internalizing the dynamic configuration of the graph. We thus expect these latent representations to express a distinct structural understanding. To investigate this, we extract the hidden states generated at inference time by both paradigms across each transformer layer. Because our primary interest lies in how the models internalize the graph structure, we isolate our analysis to the representations of the graph description tokens ($E_{G_i}$). Specifically, we examine the ability of AR and NAR models to encode the topological depth of vertices (their distance from the center) and their capacity to encode the local ``role'' of a vertex, as conveyed through unique vocabulary identifiers (Vertex IDs).

We visualize the vertex representations (averaged over occurrences in description) by projecting them into 2D using Principal Component Analysis (PCA). We then color the resulting space to indicate the topological distance from the center and the corresponding Vertex ID. Fig. \ref{fig:zoom-vmeasure}(a) and Fig. \ref{fig:embedding_ndoeid}, \ref{fig:embedding_overall} in the Appendix illustrate the layer-wise evolution of the latent space, colored using each schema. The visualizations reveal diverging strategies, the AR model strictly preserves a structured ``depth-map'' while rapidly dismantling its initial Vertex ID clustering, confirming its need to prioritize global dependencies over local identities for complex $l^{th}$-order forward planning. In contrast, the NAR model's initial depth separation quickly dissolves, yet it preserves Vertex ID clustering much deeper into the network. This diffuse topological representation, paired with strong local identity retention, perfectly aligns with its reverse-decoding strategy, demonstrating that NAR models sidestep the need for a persistent global hierarchy by relying on explicit token identities for simple $1^{st}$-order local transitions. This local strategy is further reinforced at the first layer, as can be seen in a zoom-in illustration (Fig. \ref{fig:zoom-vmeasure}(a)). Unlike AR's topologically noisy Vertex ID clusters, NAR clusters display a clear depth gradient, indicating that bidirectional attention allows NAR models to ``pre-compute'' structural relationships immediately and bypass the AR's sequential learning bottleneck entirely.

We further evaluate these topological depth claims quantitatively by calculating the V-measure clustering scores on the original high-dimensional embeddings. Figure \ref{fig:zoom-vmeasure}(b) tracks the V-measure across model layers. As shown, the representations generated by the AR model contain significantly more information regarding the distance of vertices from the center compared to NAR models. In stark contrast, the NAR model's score remains flat and low across all layers, quantitatively confirming our visual observation that its structural representation becomes diffuse and does not rely on a persistent global distance metric.

Ultimately, this provides further evidence for the distinct planning strategies adopted by each paradigm. While AR models are forced to construct and maintain a global structural hierarchy to navigate complex forward lookahead, NAR models exploit bidirectional context to rely entirely on local identity matching, seamlessly facilitating their highly efficient reverse-decoding approach.

\begin{figure}[!h]
    \centering
    \includegraphics[width=0.8\linewidth]{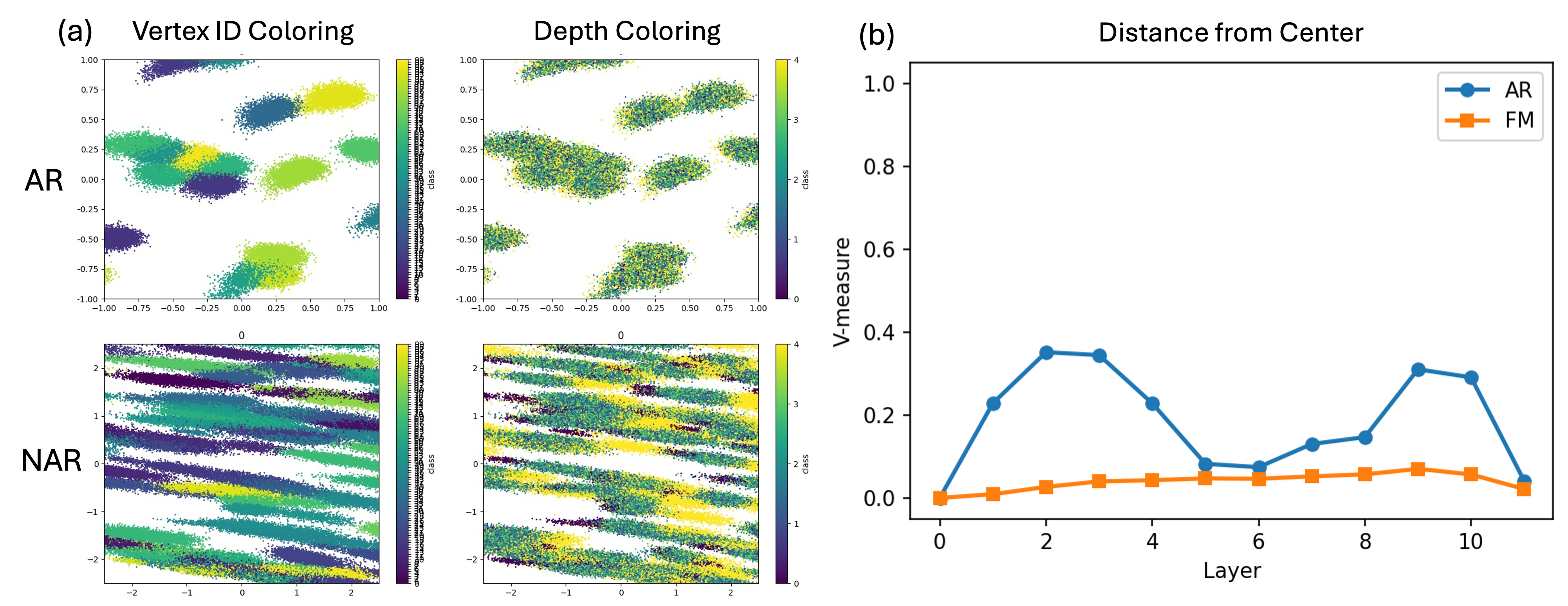}
    \caption{(a) Zoomed-in analysis of Layer 0 latent representation. While both models cluster by vertex ID, the NAR model organizes clusters into clear gradients based on topological depth. The AR model shows no such internal structure. (b) V-measure clustering scores for topological depth across model layers. The AR model actively encodes distance from the center, while the flat, low scores of the NAR model confirm it does not maintain a global structural hierarchy.}
    \label{fig:zoom-vmeasure}
\end{figure}


\section{Discussion} \label{sec:conclusion}

We investigated the internal mechanisms of AR and NAR models on lookahead planning tasks. Contrary to previous claims that AR models are structurally incapable of such planning, we demonstrate that both paradigms can achieve perfect accuracy when provided with sufficient data and a structure-enforcing objective. However, they arrive at the solution through fundamentally different computational paths. A central insight from our work is the profound impact of directionality and asymmetry as an inherent phenomenon in planning problems. While forward planning often involves navigating complex, ambiguous branching pathways, the reverse trajectory is frequently deterministic.

Constrained by causal, left-to-right decoding, AR models are forced to navigate this structural ambiguity head-on, requiring them to master complex $l^{th}$-order lookahead patterns. Our analysis of training dynamics reveals that mastering this lookahead is a strictly hierarchical learning process. AR models hit a prolonged ``learning bottleneck’’, establishing a foundation of simple $1^{st}$-order transitions before synthesizing them into global, high-order topological maps. Recognizing this inherent hierarchical dependency holds significant potential for improving the training dynamics of general LLMs.

In contrast, NAR models demonstrate remarkable generative flexibility by fully exploiting the task's structural asymmetry. By naturally adopting a ``reverse-decoding’’ strategy, they effectively bypass the high-order planning challenge altogether, reducing it to a sequence of simple $1^{st}$-order neighbor transitions. While this mechanism allows NAR models to converge with exponentially fewer training examples, it reveals that the NAR models did not actually learn to ``lookahead’’. Instead, they found a highly efficient computational shortcut. Therefore, while NAR modeling offers distinct advantages in flexibility and sample efficiency, we cannot confidently conclude that it possesses a fundamentally superior cognitive capacity for forward planning. 

Moreover, this phenomenon highlights a critical vulnerability in modern LLM training. As standard training objectives optimize for likelihood rather than true causal reasoning, we have no guarantees regarding the actual internal algorithms a model develops. Models will exploit any available asymmetry or artifact to minimize loss, often resulting in behaviors that are empirically successful but mechanically misaligned with our intentions. As advanced architectures like dLLMs become adept at finding creative shortcuts, understanding \textit{how} they solve problems becomes just as important as \textit{whether} they solve them. Rigorous investigative studies, such as this work, linking latent mechanisms to observable behaviors, are therefore not merely academic exercises but rather indispensable for ensuring reliability and safety of future AI systems.


\section*{Limitations}

While our study provides valuable mechanistic insights into how generative models approach planning tasks, our work has several limitations. First, our reliance on the minimal Star-Path framework rigorously isolates planning mechanisms, but due to its synthetic nature, it lacks the complexity, noise, and semantic richness of real-world tasks. As such, applying the conclusions in this work to actual LLMs requires further study. Secondly, because the remarkable sample efficiency of NAR models stems from exploiting structural asymmetries via reverse-decoding, it remains unclear if they hold an architectural advantage in environments with perfectly symmetrical ambiguity or where deterministic anchoring is impossible. This is especially true in the real world, where it is often difficult to identify the directionality and symmetry of problems, or whether the asymmetry is significant enough to enable a true advantage.

\section*{Disclosure of Funding}
This work was supported by Israel Science Foundation (ISF) under the Grant No. 3123/25.


\bibliographystyle{plainnat}
\bibliography{reference}


\newpage
\appendix
\onecolumn


\section{Appendix}

\subsection{AR and NAR Implementation Specifications} \label{app:architecture}

Our implementation of an autoregressive and non-autoregressive model was based on the Flow-Matching repo\footnote{\hyperlink{https://github.com/facebookresearch/flow_matching}{https://github.com/facebookresearch/flow\_matching}}. To enable comparable AR and NAR modeling, we have modified the training and inference implementations to standard autoregressive procedures based on the same architecture. The following are the model parameters used for training the AR and NAR models:

\begin{table}[ht]
\centering
\caption{Configuration hyperparameters.}
\label{tab:architecture_comparison}
\begin{tabular}{lcc}
\toprule
\textbf{Parameter} & \textbf{AR} & \textbf{NAR} \\
\midrule
Causal Masking & $\checkmark$ & $\times$ \\
Autoregressive Decoding & $\checkmark$ & $\times$ \\
\midrule
Batch Size & \multicolumn{2}{c}{1024} \\
Max. Training Examples & \multicolumn{2}{c}{$50\times10^6$} \\
\midrule
Optimizer & \multicolumn{2}{c}{AdamW} \\
Learning Rate & \multicolumn{2}{c}{$3 \times 10^{-4}$} \\
$\beta_1$ & \multicolumn{2}{c}{0.9} \\
$\beta_2$ & \multicolumn{2}{c}{0.95} \\
$\epsilon$ & \multicolumn{2}{c}{$1 \times 10^{-8}$} \\
Weight Decay & \multicolumn{2}{c}{0.03} \\
Warmup Steps & \multicolumn{2}{c}{$5 \times 10^3$} \\
Grad Clip & \multicolumn{2}{c}{1.0} \\
$\eta_{\min}$ Ratio & \multicolumn{2}{c}{0.01} \\
Accumulation Steps & \multicolumn{2}{c}{1} \\
\midrule
Loss Function & \multicolumn{2}{c}{Cross Entropy} \\
Source Distribution & Masked Seq. & $\times$ \\
Scheduler Type & Polynomial & $\times$ \\
\midrule
Hidden Size & \multicolumn{2}{c}{384} \\
Cond. Dim. & \multicolumn{2}{c}{4} \\
\# Transformer Blocks & \multicolumn{2}{c}{12} \\
\# Attention Heads & \multicolumn{2}{c}{6} \\
Dropout & \multicolumn{2}{c}{0.1} \\
\bottomrule
\end{tabular}
\end{table}

This study utilized approximately 500 GPU hours of compute time on NVIDIA L40 GPUs.


\subsection{Success Criteria} \label{app:evaluation}

In this study, we primarily focus on evaluating the performance of the model's ability to predict the path from the source to the target vertices. For this reason, we measure convergence using the \textit{exact-match} metric. A prediction is considered correct if and only if the entire predicted path matches the ground-truth path token-by-token.

Formally, let $\tilde{\boldsymbol{x}}_{\text{path}} = LM_\theta(\boldsymbol{x}_{\text{prefix}})$. Then, for a test set $\mathbb{D}$,

\begin{equation} \mathcal{A}:=\frac{\sum_{(\boldsymbol{x}_{\text{prefix}},\boldsymbol{x}_{\text{path}})\in\mathbb{D}}\mathbf{1}\!\left\{\tilde{\boldsymbol{x}}_{\text{path}}^i=\boldsymbol{x}_{\text{path}}^i \;\forall\; i\in[0,l)\right\}}{|\mathbb{D}|}
\end{equation}

where $\boldsymbol{x}^i$ denotes the $i$-th token in the sequence $\boldsymbol{x}$.

We use the validation set to evaluate \textit{exact-match} accuracy at equidistant intervals during the training process. A model is considered to have converged when the exact-match accuracy exceeds $>0.95$ for $10$ consecutive evaluation steps, or when the maximum training budget (in examples) is reached.


\subsection{Learnability of The Star-Path} \label{app:learnability}

Contrary to recent claims suggesting that AR models are fundamentally incapable of solving lookahead planning tasks \cite{bachmann2024pitfalls}, our results demonstrate that standard AR Transformers can learn the Star-Path task with perfect accuracy. 
We posit that the previously reported failures are not due to inherent structural limitations of the architecture, but rather stem from the optimization signal provided during training. 
Specifically, we identify that the learnability of the task heavily depends on whether the model is incentivized to internalize the graph structure.

\begin{figure}[!h]
    \centering
    \includegraphics[width=0.5\textwidth]{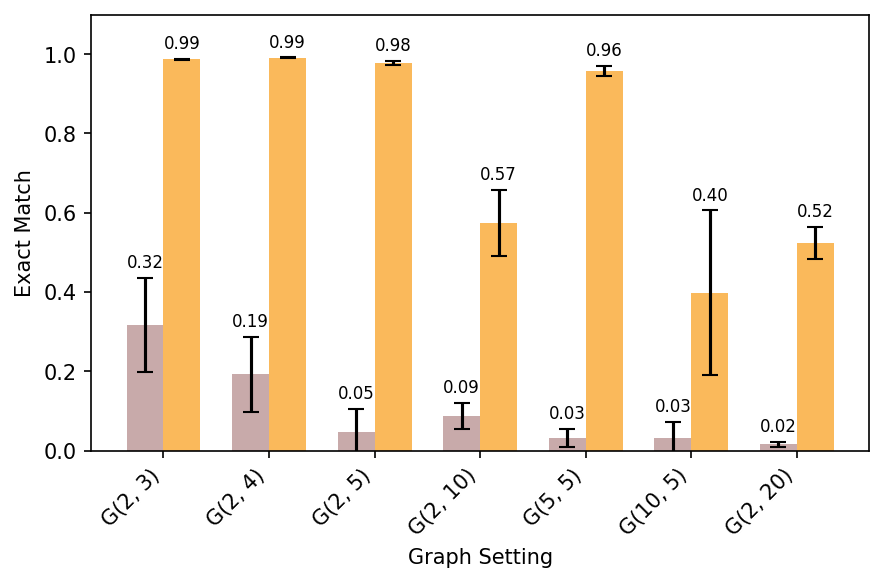}
    \caption{
        Comparison of AR models trained with (orange) and without (pink) gradients on the graph prefix across graph settings. 
        All models were trained on up to 50M distinct training examples. Confidence intervals were computed based on 10 seeded runs.
    }
    \label{fig:experiment_ar_grad_on_predix}
\end{figure}

To isolate this effect, we compare two training objectives. 
The standard approach models the conditional probability $p(\text{path}|\text{prefix})$, where the prefix (graph description and source-target pair) is provided as a prompt context without gradient propagation on its tokens. 
While theoretically sufficient, we find that this formulation renders the optimization landscape difficult to traverse, leading to slow or stalled convergence. 
In contrast, as can be seen in Fig.\ref{fig:experiment_ar_grad_on_predix}, by enabling gradient propagation on the prefix and treating the graph description as part of the generative modeling task, we observe a dramatic acceleration in convergence. 
Although the precise dynamics of this pre-training-like phenomenon warrant further investigation, it suggests that forcing the model to explicitly model the graph structure rather than merely attending to it is crucial for acquiring the necessary lookahead mechanisms.

However, we observe that even with this optimized objective, fully mastering the task necessitates a substantial volume of training data. 
In our experiments, we capped the training budget at 50M sequences, an order of magnitude more than previous work (capped at 200K with multiple epochs). 
This indicates that while learnable, the task remains sample-inefficient for AR architectures. Furthermore, we observe that for graph configurations, even this extensive budget occasionally proved insufficient. 
Crucially, all models are trained for a single epoch over non-repeating data streams. This strict regime ensures the model cannot overfit to specific training instances, further establishing that our reported results reflect generalization rather than memorization.

Fig.\ref{fig:experiment_ar_grad_on_predix} illustrates the empirical impact of this training choice across varying graph complexities. 
We compare the Exact Match accuracy of AR models trained with and without prefix gradients on an equal number of examples (50M). 
The results reveal that models trained with prefix gradients (orange bars) were able to achieve near-perfect accuracy ($>0.95$)\footnote{To reduce computational and environmental costs, runs maintaining a convergence score higher than $0.95$ for $10$ consecutive evaluation steps were terminated early.} in most settings. 
Conversely, the baseline approach (pink bars) fails to converge. 
These findings demonstrate that the primary bottleneck in AR planning is not the search capacity, but the richness of the supervision signal provided during training.

Given the potential impact of these findings, we encourage further research to better understand the intricate dynamics of this experimental insight.

\subsection{Beyond Star-Graphs} \label{app:two_sided_star}

In the main text (\S\ref{sec:method}), we justified our focus on the minimal Star-Graph topology $G(d, l)$ by reasoning that any arbitrary graph structure containing branching junctions can be fundamentally represented as a composition of simple star graphs. The minimal setting allowed us to isolate and study the learned lookahead mechanisms with maximum precision. However, to rigorously demonstrate the generalization of our findings beyond a single junction, we introduce and analyze an extension of the Star-Path, the ``Two-Sided-Star'' use-case.

\begin{figure}[!h]
    \centering
    \includegraphics[width=\textwidth]{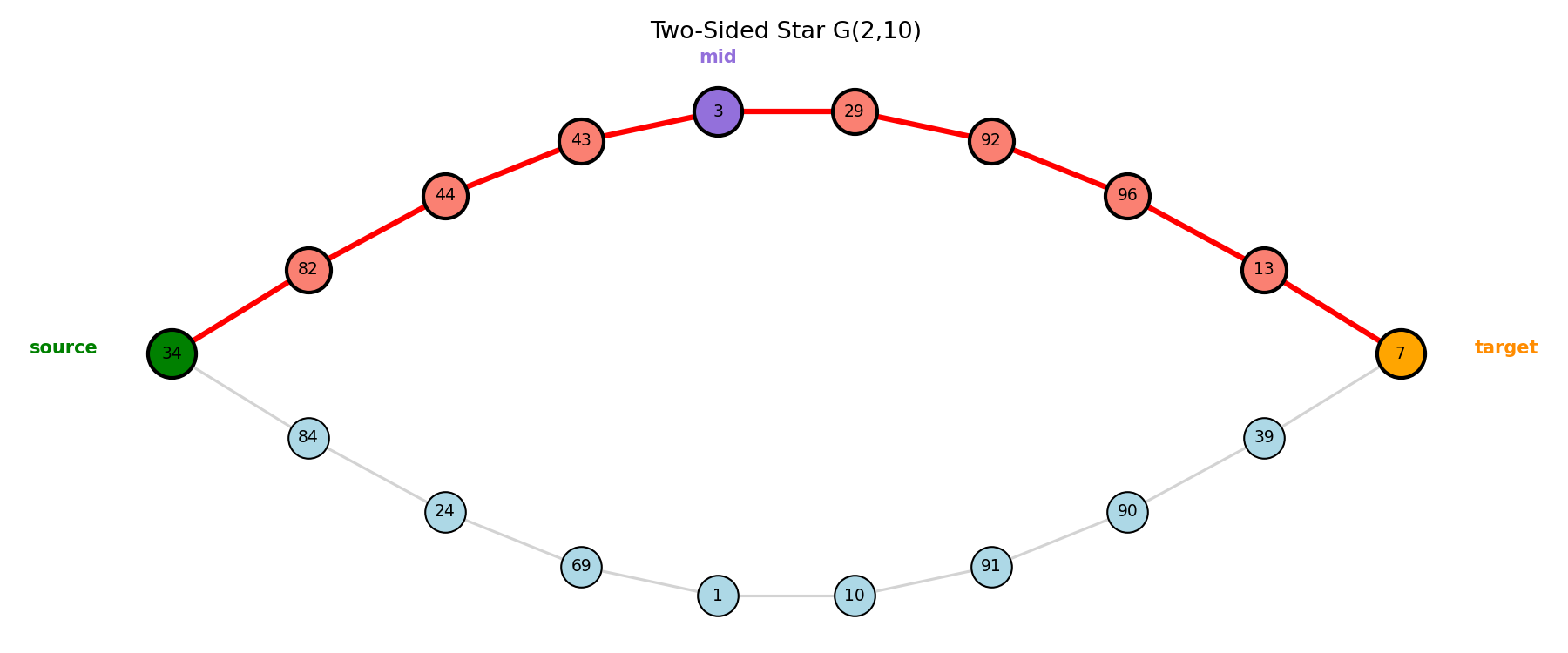}
    \caption{Structure of the ``Two-Sided-Star'' graph topology. Branching junctions exist adjacent to both the source (green) and target (orange) vertices, removing any trivial reverse-decoding advantage from the goal. The highlighted path (red) must be deduced, connecting source to target through a provided mid-path vertex (purple).}
    \label{fig:two_sided_star_example}
\end{figure}

The Two-Sided-Star configuration is designed to eliminate any inherent directional bias in the task setup. In the original Star-Path task (\S\ref{sec:star_path}), the reverse direction (from leaf to center) is naturally junction-free. To remove this asymmetry, we construct the graph structure illustrated in Fig. \ref{fig:two_sided_star_example}. This topology consists of two symmetric star-graph components (a ``source-side'' and a ``target-side'') joined by a shared central mid-path vertex ($v_{mid}$, introduced to make the path unique, shaded purple in the figure). The source vertex ($v_s$) and the target vertex ($v_t$) are placed as leaf nodes on opposing sides. Crucially, as Fig. \ref{fig:two_sided_star_example} highlights, both $v_s$ and $v_t$ face branching junctions immediately upon traversal (vertices 34 and 7, respectively). By randomizing which arm on each side completes the full path, we create a task where both the forward and backward directions from the endpoints require resolving high-order lookahead to identify the unique connecting path through $v_{mid}$. Local information alone is insufficient from either endpoint.

\begin{figure}[!h]
    \centering
    \includegraphics[width=\textwidth]{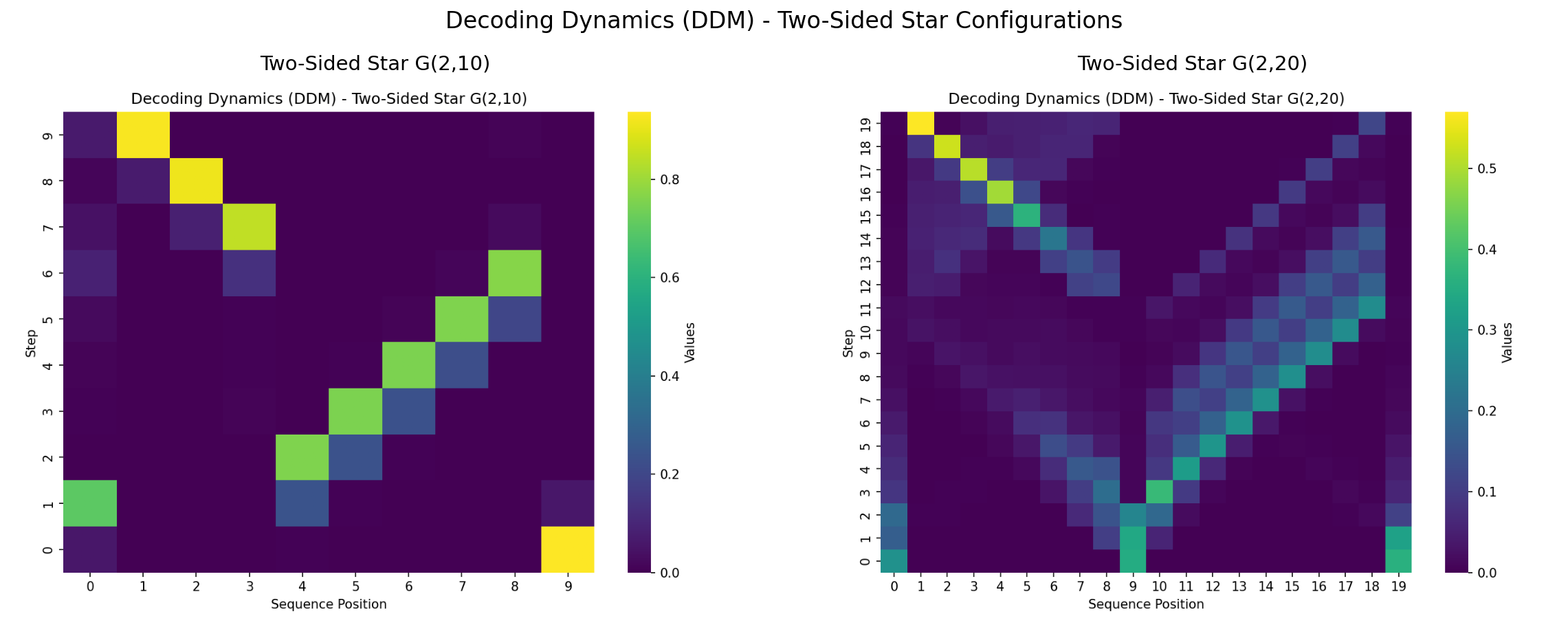}
    \caption{Decoding dynamics of a trained NAR model (dLLM) on Two-Sided-Star configurations. As unmasking proceeds over time (y-axis), the model prioritizes unmasking tokens moving outwards from the mid-path vertex towards both the target and source simultaneously. This adaptive, bidirectional expansion allows the model to solve the task using only simple $1^{st}$-order transitions, despite junctions being present from both ends.}
    \label{fig:decoding_dynamics_2_10_2_20}
\end{figure}

The expected behavior for a model truly exploiting structural asymmetry would be to identify the most deterministic part of the task and utilize that as a decoding anchor. In the Two-Sided-Star, while the endpoints are ambiguous, the path leading from the provided mid-path vertex ($v_{mid}$) outwards is deterministic. An optimal adaptive strategy would involve ``anchoring'' at $v_{mid}$ and decoding sequentially towards both $v_s$ and $v_t$ using simple $1^{st}$-order transition matching, which is junction-free. To evaluate whether NAR models adopt this expected behavior, we trained a NAR model (dLLM) on this specific composite structure. Upon evaluating the trained model's performance, we found that it converges perfectly and exhibits the sample efficiency advantages observed in the single-junction case. We explicitly examined the decoding pattern of this model to uncover its underlying strategy. The result of this mechanistic deep-dive is presented in Fig. \ref{fig:decoding_dynamics_2_10_2_20}. The visualization confirms that the model implements the precisely hypothesized dynamic adaptive decoding mechanism. Across different configurations ($G(2,10)$ and $G(2,20)$), a strikingly consistent decoding pattern emerges.

The diagonal unmasking pattern indicates that the NAR model does not decode causally (left-to-right) or strictly backward (right-to-left). Instead, it dynamically expands outwards from the provided mid-path vertex towards both sides. This strategy brilliantly reduces the complex global planning problem into two parallel sets of simple, $1^{st}$-order deterministic neighbor transitions. This is functionally equivalent to the adaptive mechanism discovered in the original star-path, but here applied to a composite structure where simple reverse-decoding from the target was made impossible. This finding strongly suggests that Discrete Diffusion models are not exploiting a simple directionality trick, but are dynamically adapting their generation order to find the most efficient computation path, a capability fundamentally unavailable to Autoregressive models forced into causal sequential constraints.


\subsection{Mechanisitc Adaptability - Additional Results} \label{app:mech_additional}

\begin{figure}[!h]
    \centering
    \includegraphics[width=0.5\textwidth]{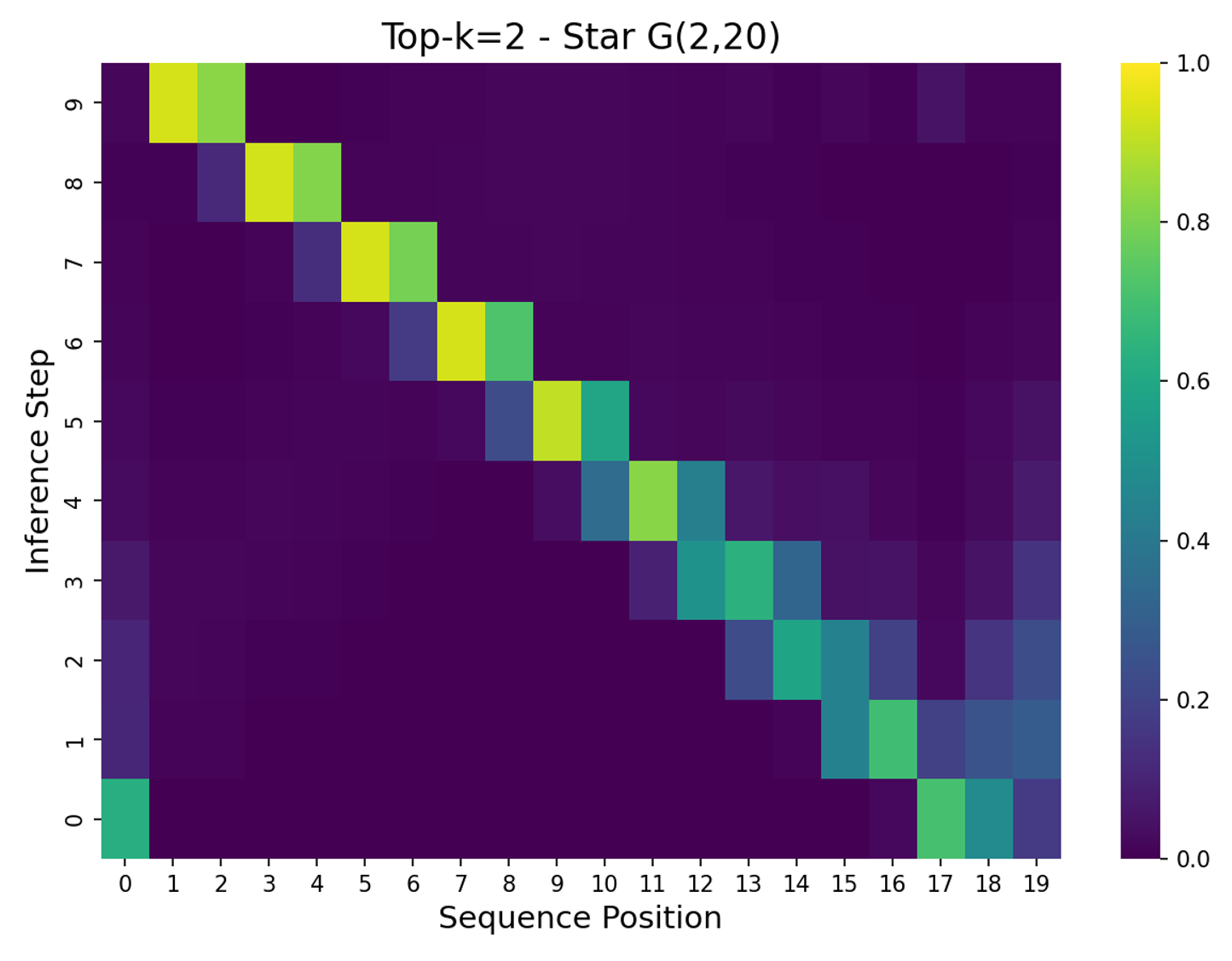}
    \caption{Decoding dynamics for $k=2$ on the $G(2,20)$ graph. Even when unmasking multiple tokens simultaneously (top-$k$ logit strategy), the model consistently utilizes $1^{st}$-order connections in a reverse, non-autoregressive manner to solve the task, mirroring the baseline $k=1$ strategy.}
    \label{fig:inference_dynamics_k2}
\end{figure}

\subsubsection{Accelerated Sampling} \label{app:mech_additional:acce}

In this work, we initially focused on the standard inference setup for discrete diffusion models, unmasking a single token per inference step ($k=1$). However, a major motivation for utilizing dLLMs is their capacity to significantly improve decoding speed by unmasking multiple tokens ($k>1$) at each decoding step. To evaluate whether the models maintain their robustness and structural problem-solving behavior under these faster sampling strategies, we investigate the decoding dynamics when unmasking $k > 1$ tokens per step. Specifically, we apply a commonly used unmasking strategy where the top-$k$ tokens are selected based on their highest predicted logit values. All resulting figures and metrics are computed over the validation set of size 20K graph configurations.

The results demonstrate that for smaller values of $k$, the model's fundamental strategy remains completely intact. Fig. \ref{fig:inference_dynamics_k2} illustrates the decoding pattern for $k=2$ on the $G(2,20)$ topology. The visualization shows that even when unmasking multiple tokens simultaneously, the model consistently utilizes simple $1^{st}$-order connections in a non-autoregressive manner to solve the task. By successfully anchoring at the target and decoding backward, the model's behavior perfectly mirrors the $k=1$ baseline.

\begin{figure}[!h]
    \centering
    \includegraphics[width=\textwidth]{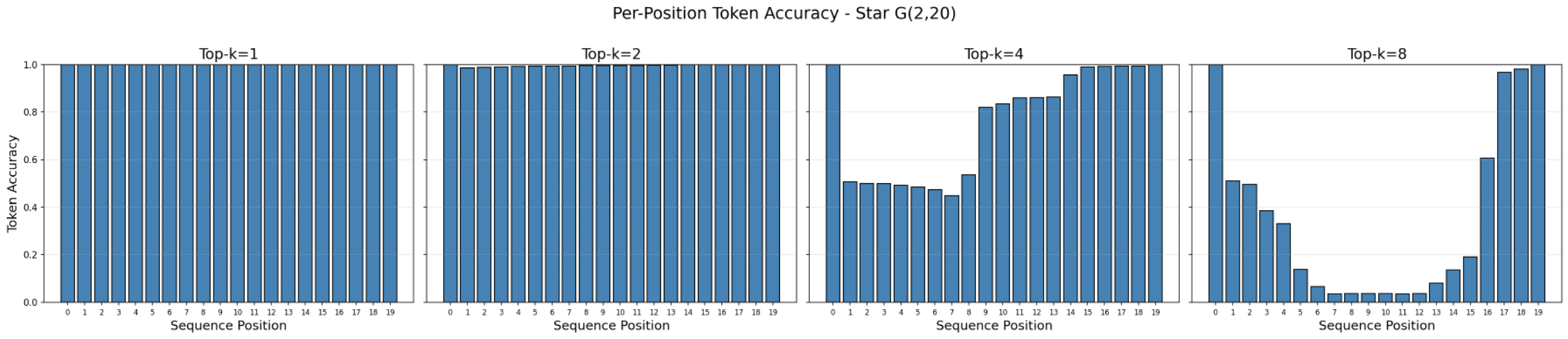}
    \caption{Decoding dynamics for $k=4$ (left) and $k=8$ (right) on the $G(2,20)$ graph. As the number of simultaneously unmasked tokens increases, the strictly sequential reverse-decoding strategy breaks down, and the model prematurely attempts forward, higher-order predictions.}
    \label{fig:inference_dynamics_topk_accuracy}
\end{figure}

\begin{figure}[!h]
    \centering
    \includegraphics[width=\textwidth]{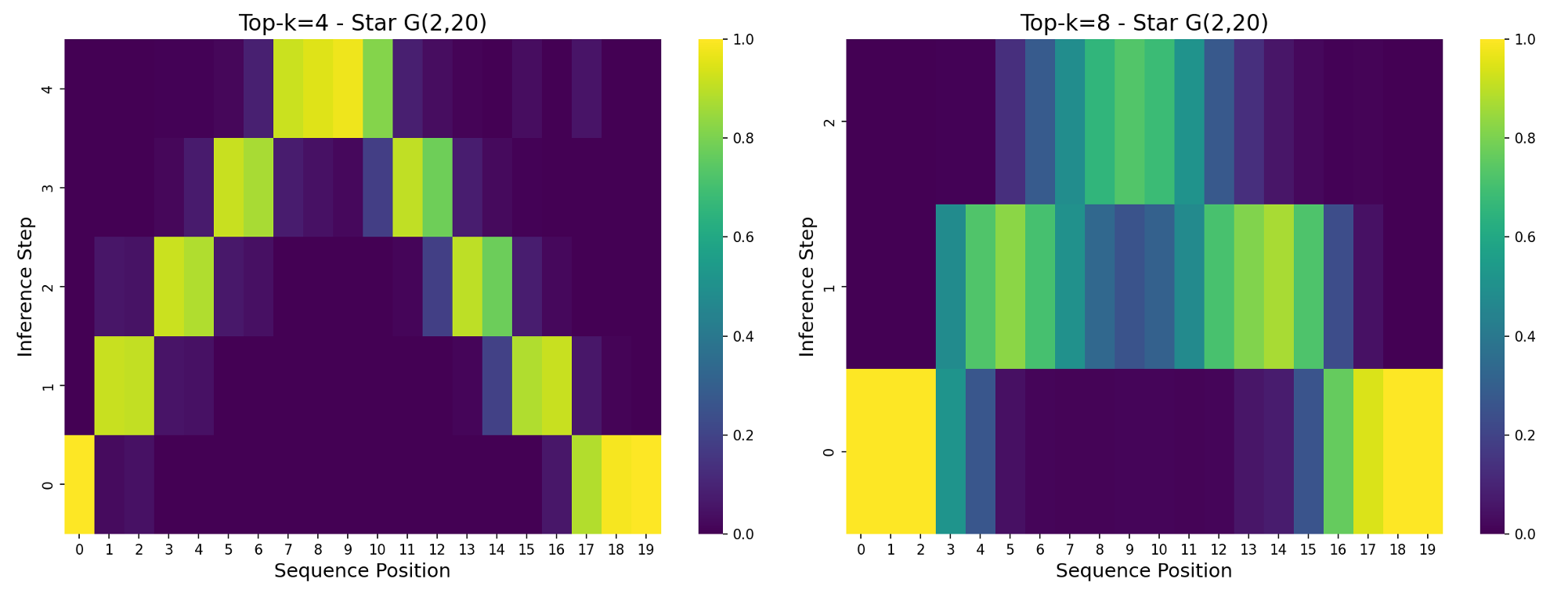}
    \caption{Mean per-position token accuracy across different top-$k$ unmasking configurations on the $G(2,20)$ test set.}
    \label{fig:inference_dynamics_topk}
\end{figure}

However, as we push the limits of this accelerated sampling by increasing $k$ to 4 and 8 (Fig. \ref{fig:inference_dynamics_topk_accuracy}), this rigid reverse-decoding pattern begins to fracture. While the model still attempts to decode backwards from the goal, it simultaneously begins unmasking tokens in the forward direction. This dual-directional unmasking is accompanied by a significant degradation in path accuracy. As depicted in Fig. \ref{fig:inference_dynamics_topk}, while the $k=1$ and $k=2$ settings maintain near-perfect mean per-position accuracies of 0.9996 (min 0.999) and 0.9957 (min 0.987), respectively, increasing $k$ leads to a severe performance drop. For $k=4$, the overall accuracy falls to 0.7555, with intermediate sequence positions dropping to $\sim 0.45-0.50$ before recovering toward the end of the path. Under the extreme $k=8$ setting, the model struggles drastically, achieving a mere 0.3555 accuracy. At this accelerated rate, the intermediate path positions plummet to an accuracy of roughly 0.04, successfully resolving only the immediate adjacent vertices (positions 0 and 19) alongside the tail end of the path, further demonstrating that local $1^{st}$-order transitions were successfully learned. Crucially, because the model fundamentally fails to solve the task at these higher $k$ values, these fractured decoding patterns cannot be reliably used to either support or contradict our core claims regarding lookahead mechanisms. Nevertheless, the specific nature of this failure provides an interesting insight into the model's internal representations.

This breakdown reveals an intriguing nuance regarding the internal representations learned by the NAR model. The fact that the model attempts forward prediction for high $k$ values, albeit unsuccessfully, implies that it did acquire some latent representation of higher-order lookahead ($>1$) during training. However, this high-order understanding is fragile. Not only does it struggle to resolve higher-order transitions from the forward direction, but it also has a hard time accurately following 4 or 8 transitions backward in a single step. Without the step-by-step, deterministic grounding provided by sequential $1^{st}$-order reverse transitions, the model's internal representations are insufficiently precise to accurately bridge the long-distance dependencies required to solve the task.

We leave further investigation to future work.

\subsubsection{Block-wise decoding} \label{app:mech_additional:block}

\begin{figure}[!h]
    \centering
    \includegraphics[width=0.5\textwidth]{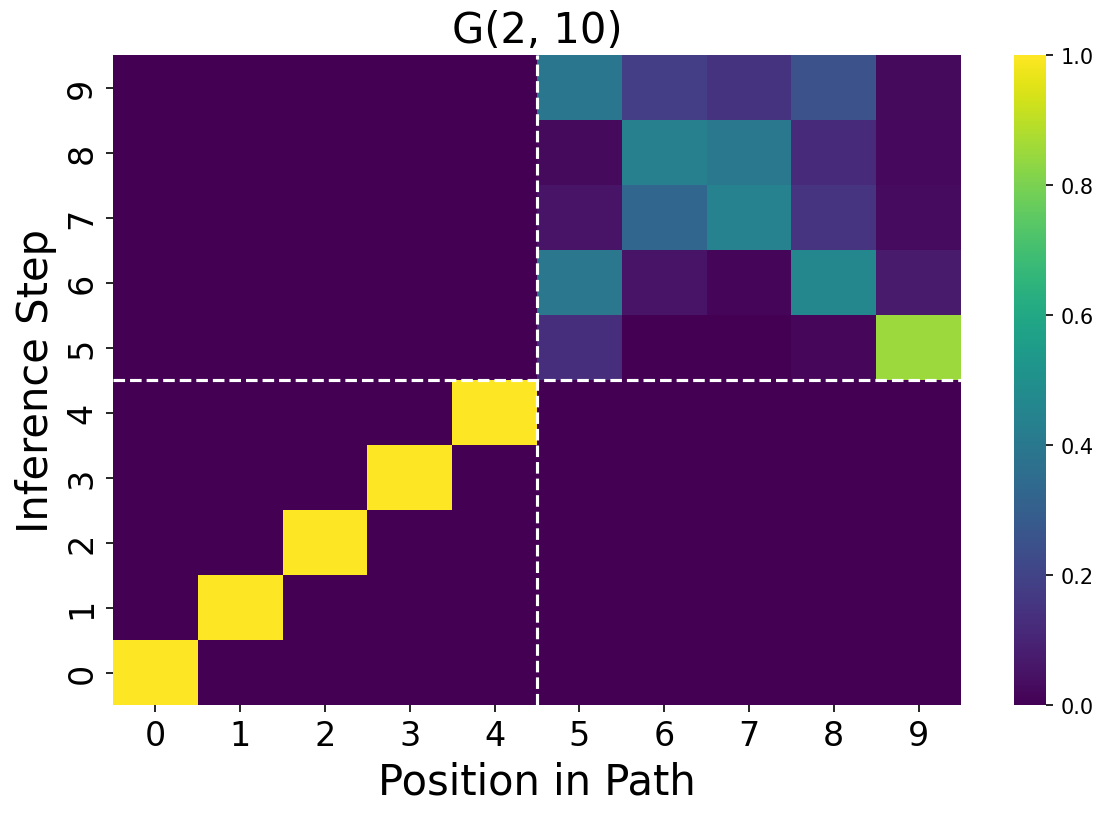}
    \caption{Block-wise decoding dynamics of the NAR model on \(G(2,10)\) using two generation blocks of five path tokens each. Dashed lines indicate the block boundary. Color intensity indicates the fraction of examples in which each path position is unmasked at each inference step. During the first block, target-side information is unavailable and the model decodes from left to right. Once the second block becomes active and the target is accessible, the model anchors at the final path position and decodes predominantly backward within the block.}
    \label{fig:blockwise_decoding}
\end{figure}

The experiments above use a single generation block, allowing the model to condition on partially decoded tokens across the entire output sequence. In practical settings, however, outputs longer than the generation block must be produced block-wise, preventing later blocks from providing information while earlier blocks are generated. To examine how this constraint affects the learned strategy, we evaluate the NAR model on \(G(2,10)\) using two blocks of five path tokens each (Fig.~\ref{fig:blockwise_decoding}). During generation of the first block, target-side information is unavailable, and the model therefore decodes the block in the forward direction. Once generation enters the second block and the target becomes available, the decoding order changes, the model anchors at the final path position and proceeds predominantly backward through the block. Thus, block-wise generation prevents the model from exploiting the global reverse-decoding shortcut across block boundaries, but the model still adaptively exploits deterministic reverse transitions whenever the required information becomes available.

\clearpage

\subsection{Training Dynamics} \label{app:training_dynamics}

\begin{figure}[!h]
    \centering
    \includegraphics[width=0.75\linewidth]{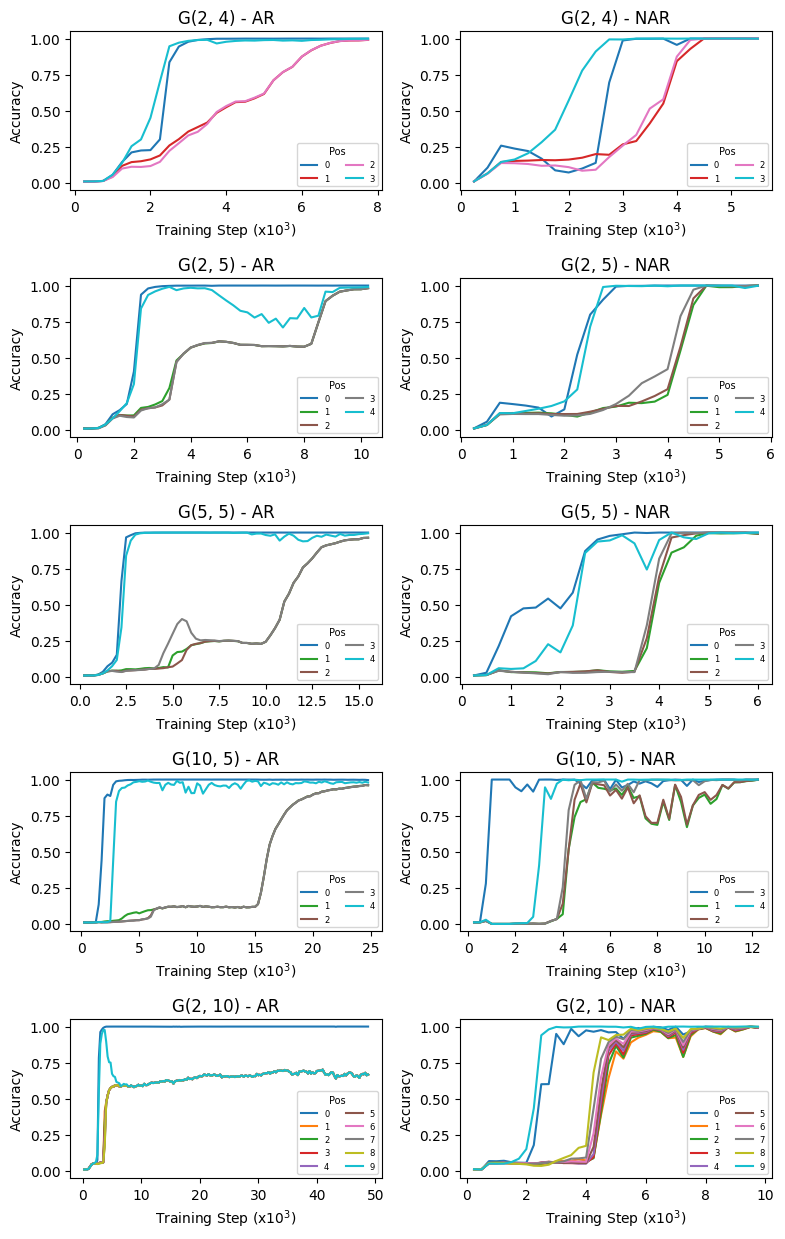}
    \caption{A comparison of the training dynamics of AR and NAR models depicting the per-position (vertex) average accuracy over the validation set. Note that ``\textit{G(2,10) - AR}'' did not reach convergence. For visualization purposes, the x-axis scale varies across.}
    \label{fig:training_dynamics}
\end{figure}

We further examine the convergence of individual token positions (Fig. \ref{fig:training_dynamics}, Appendix \ref{app:training_dynamics}). While both paradigms quickly learn to copy the path endpoints, their mastery of intermediate vertices differs fundamentally. The AR model exhibits a distinct ``bottleneck'', struggling to predict the intermediate path before all middle vertices emerge simultaneously. Conversely, the NAR model learns incrementally, converging on correct predictions backward from the path's end. Freed from strict sequential dependencies, it capitalizes on local $1^{\text{st}}$-order transitions while the AR counterpart must globally resolve the overarching $\ell^{\text{th}}$-order dependencies. By solving these easier sub-tasks and using converged vertices as a scratchpad for context, the NAR model implements an efficient alternative that bypasses the AR bottleneck entirely.

\subsection{Robustness to Model Scale} \label{app:model_scale}

\begin{figure}[!ht]
    \centering
    \includegraphics[width=0.7\textwidth]{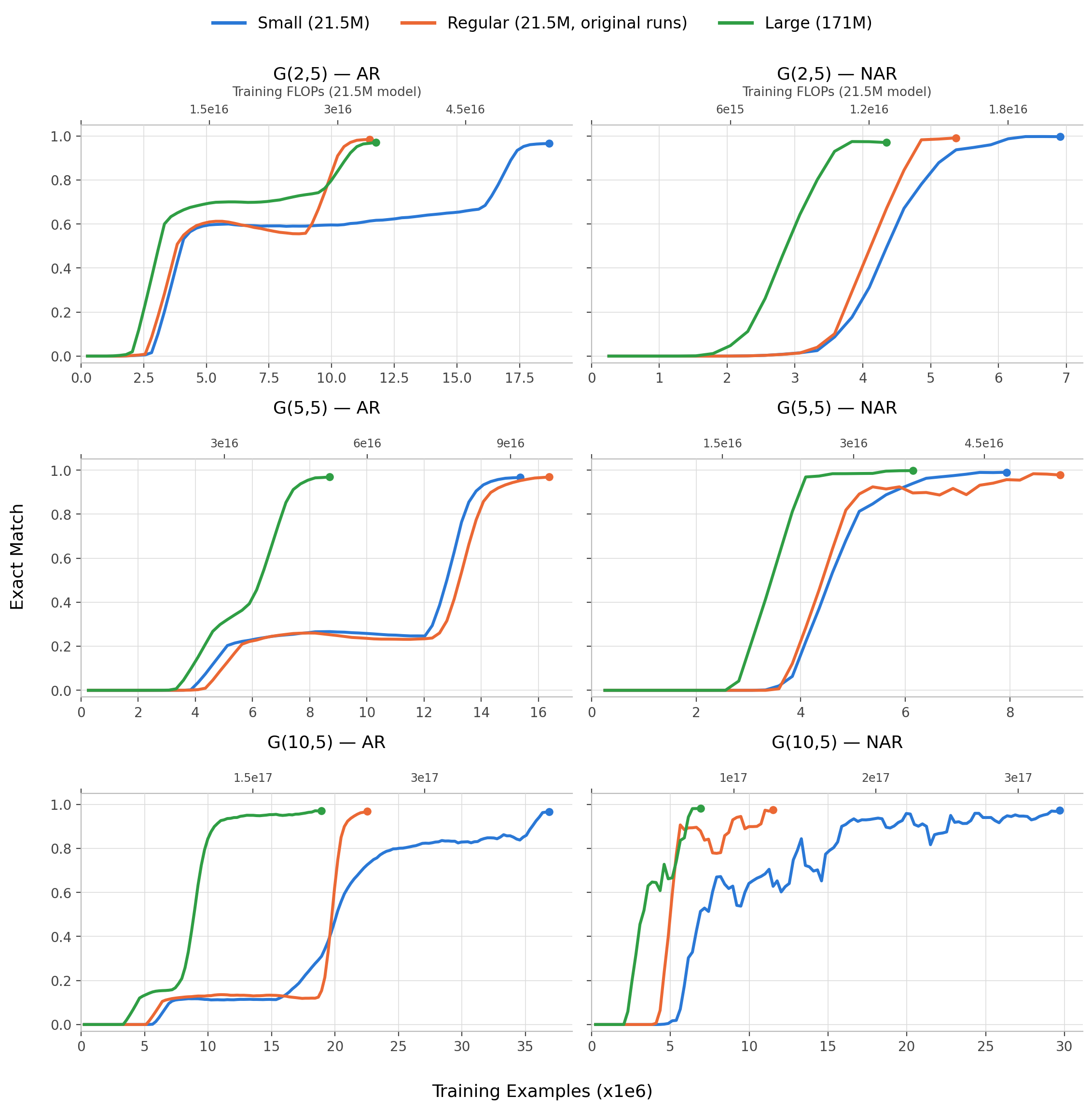}
    \caption{Robustness to model scale. Across smaller and larger model variants, NAR retains its sample-efficiency advantage over AR, while the characteristic difference in their learning dynamics remains qualitatively unchanged.}
    \label{fig:model_scale}
\end{figure}

The main experiments use a fixed Transformer model size, raising the possibility that the observed difference between AR and NAR training dynamics is specific to that capacity. To test this, we repeat the original Star-Path experiments across multiple model scales, while keeping the task and training setup otherwise unchanged. Figure
\ref{fig:model_scale} shows the exact-match trajectories for $G(2,5)$, $G(5,5)$, and $G(10,5)$ for both AR and NAR models. The original 21.5M-parameter runs are included as a reference.

Across model scales, the central qualitative behavior is preserved. Increasing model capacity generally reduces the number of training examples required to reach high accuracy, but it does not eliminate the
difference between the two generative paradigms. NAR models consistently reach high exact-match accuracy substantially earlier in training, whereas AR models retain the characteristic intermediate regime before convergence. This distinction becomes particularly clear as graph complexity increases: on $G(5,5)$ and $G(10,5)$, increasing AR model capacity shifts the final transition to earlier in training, but does not remove the separation between the initial acquisition of simpler transitions and the later acquisition of the full lookahead solution.

These results further support our interpretation of the sample-complexity gap as a consequence of the different solution strategies available to AR and NAR models, rather than an artifact of insufficient model capacity. Additional capacity can make the higher-order computation easier to acquire, but it does not change the causal constraint faced by AR models: the correct forward transition must still be resolved before future path tokens are generated. In contrast, NAR models retain the ability to exploit bidirectional context and construct the solution
from deterministic local transitions. Thus, model scale changes when the relevant behaviors are learned, but does not change the qualitative distinction in how the two paradigms learn to solve the task.

\clearpage

\subsection{Comparing Alternative AR Architectures} \label{app:mamba}

To further isolate the source of the autoregressive behavior observed in the main experiments, we also evaluate a Mamba-based autoregressive model \citep{gu2024mamba,dao2024transformers,lahoti2026mamba3}. Since Mamba remains causally constrained, it cannot exploit the reverse-decoding strategy available to the NAR model, but it replaces the Transformer backbone with a recurrent state-space architecture. Thus, it allows us to separate effects that stem from autoregressive decoding itself from effects that may be specific to the Transformer implementation of AR modeling. We train the Mamba under the same data and optimization setup as the other models (See Table \ref{tab:mamba-hyperparams}).

\begin{wrapfigure}{r}{0.5\textwidth}
    \centering
    \includegraphics[width=0.5\textwidth]{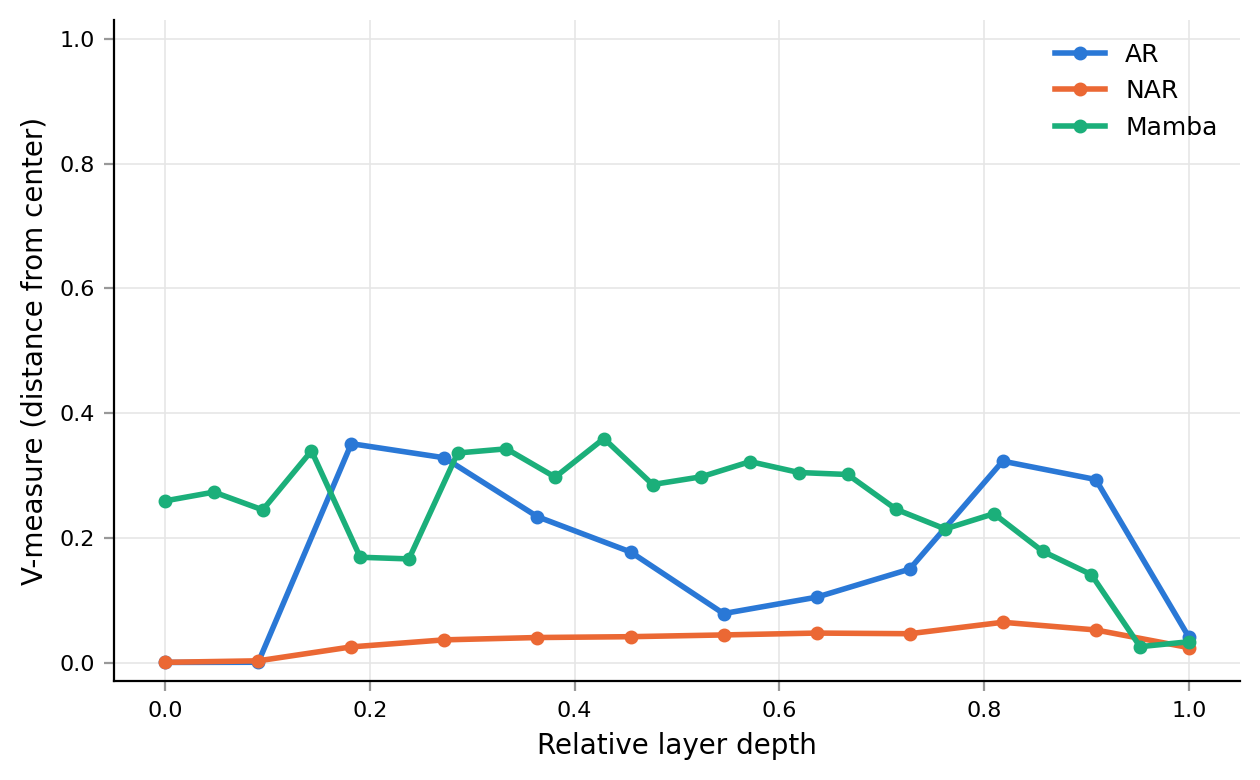}
    \caption{V-measure of latent representations with respect to topological distance from the center, plotted across relative layer depth for Transformer AR, NAR, and Mamba. Mamba maintains a stronger and more persistent structural signal than NAR, and in many layers stronger than Transformer AR, suggesting it encodes graph topology more effectively from the graph-description prefix.}
    \label{fig:vmeasure_mamba}
\end{wrapfigure}

\begin{figure}[!ht]
    \centering
    \includegraphics[width=\textwidth]{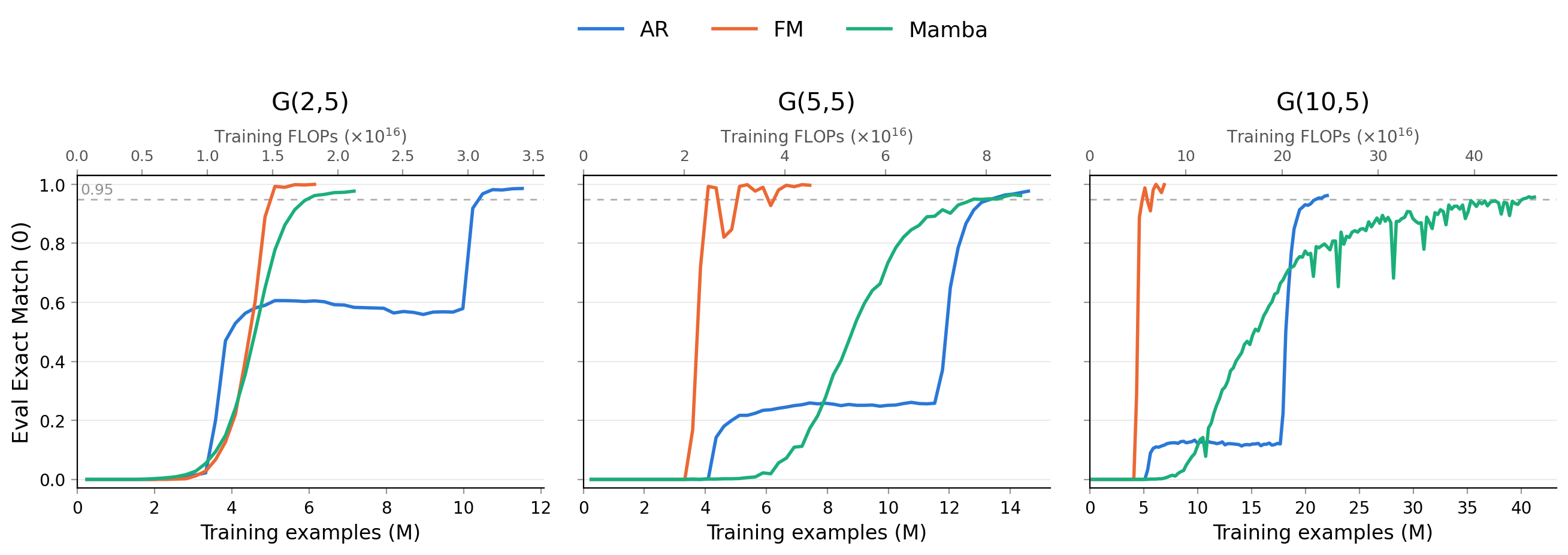}
    \caption{Training dynamics of Transformer AR, NAR (dLLMs), and Mamba AR models across increasingly complex Star-Path configurations. Mamba exhibits smoother learning than Transformer AR, but NAR retains a substantial sample-efficiency advantage as graph complexity increases.}
    \label{fig:mamba}
\end{figure}

Figure \ref{fig:mamba} exhibits substantially smoother training dynamics for the Mamba architecture, suggesting that its internal computation of the graph structure differs from the one learned by Transformer self-attention. To probe this more directly, we measure the V-measure of representations with respect to topological distance from the center, following the same structural criterion used in \S\ref{sec:embeds}. As shown in Fig. \ref{fig:vmeasure_mamba}, Mamba maintains a substantially stronger distance-from-center signal across depth than NAR, and in many layers also stronger than Transformer AR. This suggests that Mamba constructs a more persistent representation of graph structure from the graph description alone.

At the same time, Mamba does not remove the core limitation of autoregressive decoding, as unlike NAR, it still cannot adapt generation order to the structure of the task. Thus, even if Mamba represents the graph more effectively than Transformer AR, it does not gain the reverse-decoding shortcut that underlies the NAR sample-efficiency advantage.

Overall, these findings refine our interpretation of the main results. The Mamba results suggests that alternative AR backbones may learn smoother and structurally richer internal representations than Transformer AR, although limited by causal decoding. A fuller mechanistic analysis of how Mamba stores and composes graph structure is left for future work.

\begin{table}[t]
\centering
\begin{tabular}{lc|lc}
\hline\hline
\textbf{Parameter} & \textbf{Mamba} & \textbf{Parameter} & \textbf{Mamba} \\
\hline
Batch Size                & 1024               & Causal Masking          & \checkmark \\
Max.\ Training Examples   & $50 \times 10^{6}$ & Autoregressive Decoding & \checkmark \\
\hline
Optimizer                 & AdamW              & Loss Function           & Cross Entropy \\
Learning Rate             & $3 \times 10^{-4}$ & Source Distribution     & Masked Seq. \\
$\beta_1$                 & 0.9                & Scheduler Type          & Polynomial \\
\cline{3-4}
$\beta_2$                 & 0.95               & Hidden Size             & 384 \\
$\epsilon$                & $1 \times 10^{-8}$ & Cond.\ Dim.             & $\times$ \\
Weight Decay              & 0.03               & \# Mamba Blocks         & 22 \\
Warmup Steps              & $5 \times 10^{3}$  & State Size              & 16 \\
Grad Clip                 & 1.0                & Expansion Factor        & 2 \\
$\eta_{\min}$ Ratio       & 0.01               & Conv.\ Kernel Size      & 4 \\
Accumulation Steps        & 1                  & Dropout                 & 0.0 \\
\hline\hline
\end{tabular}
\caption{Hyperparameters for the Mamba baseline. Training hyperparameters are matched to the AR
and NAR models in Table~\ref{tab:architecture_comparison}.}
\label{tab:mamba-hyperparams}
\end{table}


\subsection{Generalization Beyond Star-Path} \label{app:real_planning}

Our use of the synthetic Star-Path task coupled with dedicated training deliberately isolates the effect of the generative paradigm on lookahead and enables control over the cause of the effect. However, as such, it abstracts away some aspects of practical use cases. We therefore complement the mechanistic study with an external, more confounded, study of full practical AR and NAR LLMs using established planning benchmarks. We compare two diffusion language models (DiffusionGemma-26B-A4B-it \citep{google2026diffusiongemma,gemmateam2026gemma4} and Dream-7B \citep{ye2025dream}), against three autoregressive (AR) baselines (Gemma-4-26B-A4B-it \citep{gemmateam2026gemma4}, Qwen2.5-7B \citep{qwen2024qwen25}, and Qwen3.5-9B-Base \citep{qwen2026qwen35}). Particularly, DiffusionGemma is derived from Gemma-4, and Qwen2.5-7B is not instruction-tuned and does not have a dedicated post-trained thinking/reasoning mode. Importantly, Gemma-4 and Qwen3.5 are post-trained and therefore have direct access to scratchpad, alleviating the reverse decoding advantage of NAR models as discussed in \ref{sec:intro}.  

\subsubsection{Tasks}
We evaluate on three benchmarks that require decisions whose validity depends on non-local future constraints. 

\textbf{Trip Planning} \citep{zheng2024naturalplan} \footnote{\url{https://github.com/google-deepmind/natural-plan}} is an itenerary planning dataset. Given cities to visit, required stay durations, and available direct flights, the model must produce a complete day-by-day itinerary. To solve the task efficiently, the model must therefore maintain a globally consistent trajectory rather than make independently correct local choices. We use all 1,600 test instances, spanning 3--10 cities. 

\textbf{Countdown} \citep{ye2024beyond}, a model must combine the input numbers using $+,-,\times,\div$ to reach a target value. Each arithmetic choice changes the set of reachable future states, making successful solutions depend on multi-step lookahead rather than isolated next-token prediction. We use 2,976 test tasks. 

\textbf{Sudoku} \citep{ye2024beyond}, the model is prompted to solve a given Sudoku table. We use 100 test tasks.

\subsubsection{Experimental setup}
All models are evaluated purely in context, without task-specific fine-tuning, following the few-shot protocol of Dream-7B \citep{ye2025dream}. Dream uses its published diffusion inference settings, with the number of denoising steps equal to the output-token budget: 256 for Trip Planning, 32 for Countdown, and 24 for Sudoku.

\subsubsection{Results}
\begin{figure}[t]
    \centering
    \includegraphics[width=\linewidth]{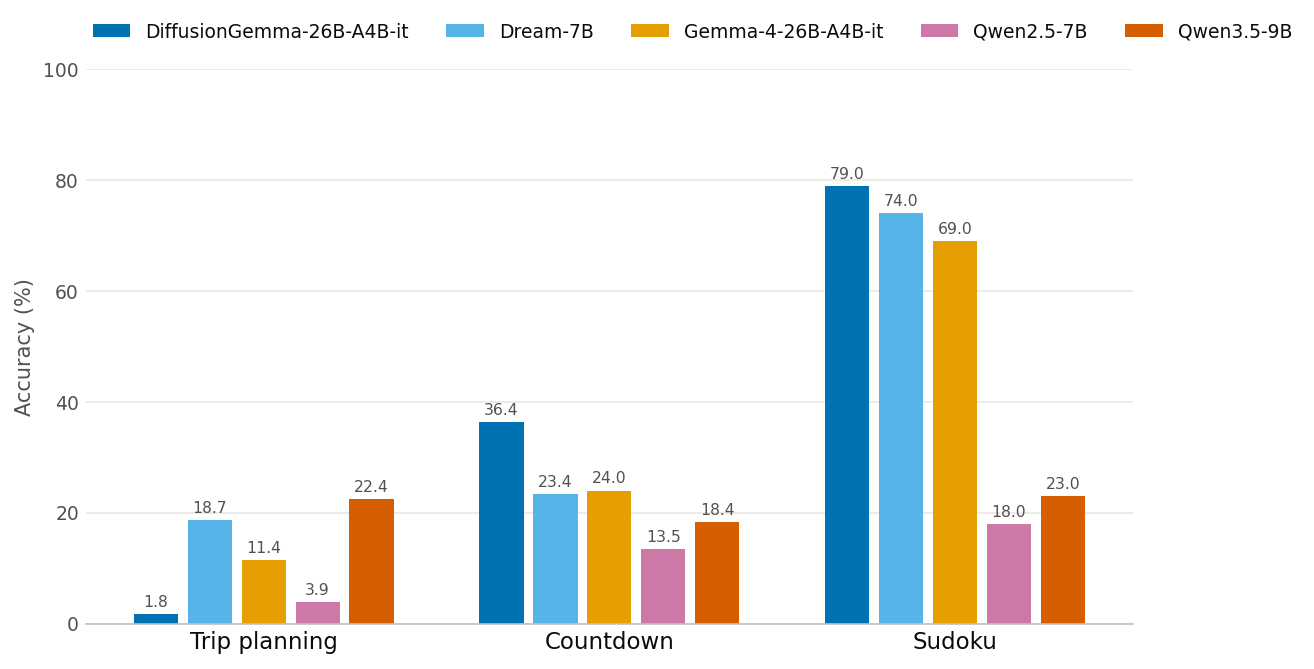}
    \caption{AR and NAR model performance on practical planning tasks using commonly used fully-trained LLMs.}
    \label{fig:real_planning}
\end{figure}

Figure~\ref{fig:real_planning} shows an overall advantage for diffusion language models, with the clearest gains on Countdown and Sudoku. Diffusion models consistently outperform the autoregressive baselines, and DiffusionGemma is the strongest model evaluated. These results further show the advantage of NAR models on lookahead tasks, and especially support the generalization of our hypothesis to practical setups. 

The only exception is observed in the Trip-Planning task where Dream diffusion model remains competitive, but DiffusionGemma performs substantially worse than the other models. A closer analysis of Trip-Planning suggests that DiffusionGemma's Trip Planning score is strongly affected by its inability to produce an output compatible with the output-parser.

Taken together, these results provide complementary behavioral support for our controlled findings. The Star-Path experiments show how non-causal generation can exploit information across the output trajectory to avoid difficult forward commitments, while the full-scale evaluation shows that diffusion models are also strong on established planning benchmarks characterized by interdependent decisions and global constraints. 

Since released checkpoints differ in model size, pretraining, post-training, and inference procedure and in particular, the Gemma variants are instruction-tuned, while Qwen2.5-7B is a base checkpoint without instruction tuning or a dedicated post-training thinking mode, we interpret these results as supporting the overall behavioral advantage of diffusion models, rather than as a causal comparison of the two generative paradigms.

\clearpage

\subsection{Latent Space Implications - Additional Results} \label{app:implications}

\begin{figure*}[ht]
    \centering
    \includegraphics[width=\textwidth]{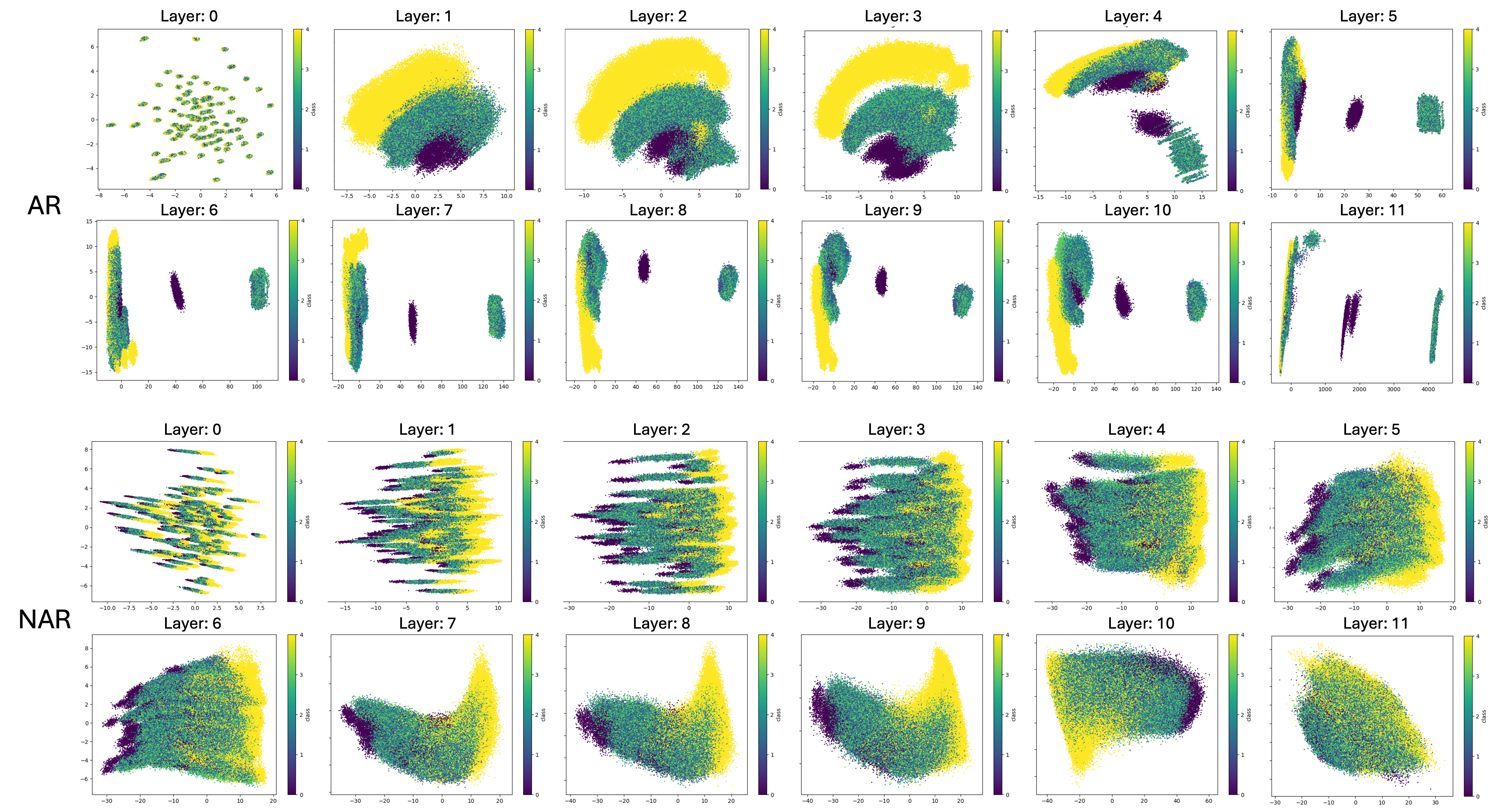}
    \caption{
        PCA projection of latent representations across layers for AR (top) and NAR (bottom) models on $G(5,5)$. 
        Points are colored by topological depth (Purple: Center, Yellow: Leaf). 
        While AR maintains structural separation deep into the network, NAR's structural representation becomes diffuse in later layers.
    }
    \label{fig:embedding_overall}
\end{figure*}

\begin{figure*}[ht]
    \centering
    \includegraphics[width=\textwidth]{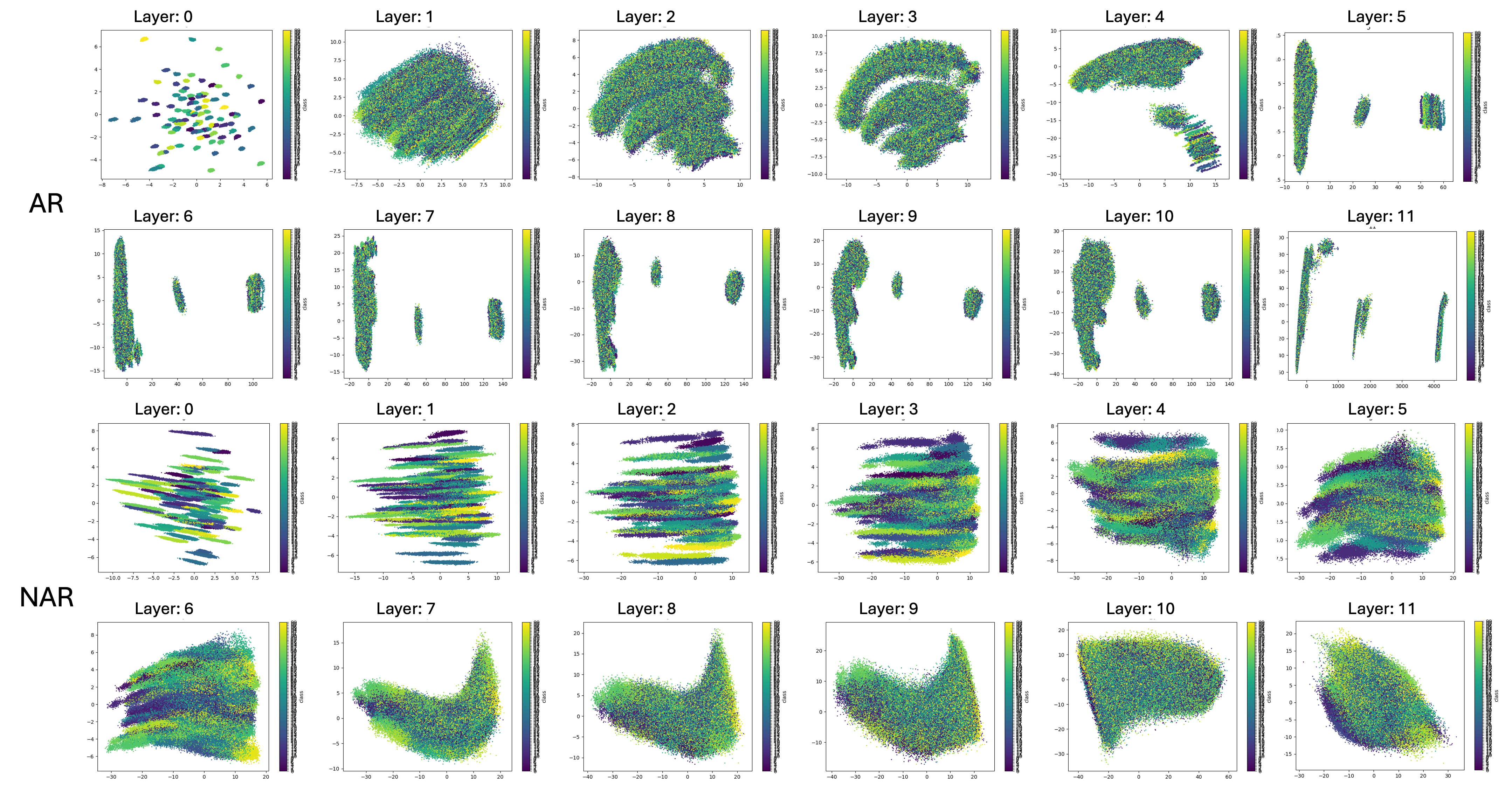}
    \caption{
        PCA projection of latent representations colored by vertex ID. NAR models (bottom) preserve identity-based clustering deeper into the network compared to AR models (top), suggesting a reliance on local identity matching.
    }
    \label{fig:embedding_ndoeid}
\end{figure*}


\newpage

\end{document}